# REGULATORY MARKETS: THE FUTURE OF AI GOVERNANCE

**Gillian K. Hadfield**[*]
**Jack Clark**[**]

**ABSTRACT:** Appropriately regulating artificial intelligence (AI) is an increasingly urgent and widespread policy challenge. We identify two primary, competing problems. First is a technical deficit: Legislatures and regulators face significant challenges in rapidly translating conventional command-and-control legal requirements into technical requirements. Second is a democratic deficit: Overreliance on industry to provide technical standards fails to ensure that the many values-based decisions that must be made to shape AI development and deployment are made by democratically accountable public, not private, actors. We propose a solution: regulatory markets, in which governments require the targets of regulation to purchase regulatory services from a government-licensed private regulator. This approach to AI regulation could overcome the limitations of both command-and-control regulation and excessive delegation to industry. Regulatory markets could enable governments to establish policy priorities for the regulation of AI, while relying on market forces and industry R&D efforts to pioneer the technical methods of regulation that best achieve policymakers' stated objectives.

Recent years have witnessed a flood of striking achievements in artificial intelligence (AI). The year 2012 saw sudden advances in image classification.[1] Five years later, deep reinforcement learning demonstrated unexpected capacities in narrow tasks like the game of *Go*.[2] Coming into public view with the release of ChatGPT in November 2022, large generative models, with billions or even trillions of parameters and trained on billions of words and images have shown surprising foundational capacities to write plausible text or computer code, illustrate ideas, generate images or video on command, answer complex questions, and much more.[3]

---

[*]Johns Hopkins University and Vector Institute.
[**]Anthropic.

1. *See, e.g.*, Alex Krizhevsky et al., *ImageNet Classification with Deep Convolutional Neural Networks*, *in* ADVANCES IN NEURAL INFORMATION PROCESSING SYSTEMS 25: 26TH ANNUAL CONFERENCE ON NEURAL INFORMATION PROCESSING SYSTEMS 2012, at 1097 (2012).
2. *See* David Silver et al., *Mastering the Game of Go with Deep Neural Networks and Tree Search*, 529 NATURE 484 (2016).
3. *See, e.g.*, Ian Goodfellow et al., *Generative Adversarial Networks*, COMMC'NS ACM, Nov. 2020, at 139; Jason Wei et al., *Emergent Abilities of Large Language Models*, 1, 6 ARXIV (Aug. 2022), https://arxiv.org/pdf/2206.07682 [https://web.archive.org/web/20250511211434/https://arxiv.org/pdf/2206.07682]; Junping Zhang et al., *Recent Advances in Artificial Intelligence Generated Content*, 25 FRONTIERS INFO. TECH. & ELEC. ENG'G 1 (2024).

The past decade has also witnessed an evolution in the conversation around AI governance. Early recognition that opaque classification systems trained on casually chosen data (images or words scraped from the internet, a history of hiring decisions) will reproduce and possibly amplify racial and gender biases sparked global conversations around AI ethics.[4] Soon a flurry of principles and guidelines emerged from industry and civil society organizations.[5] Now, governments have begun to explore—and in some cases pass—legislation to govern the development and use of a technology that is already embedded in daily life through AI-powered internet platforms and devices.

The challenge of governing AI is, however, enormous. Any new technology is faced with the “pacing” problem[6] —the lag between innovation at the speed of industry labs and governance at the speed of politics and bureaucracies—and AI is no exception.[7] But the challenge of governing AI goes beyond delay to a fundamental mismatch between the capacity of traditional governance tools and the nature of the solutions needed to ensure that a technology that is likely to touch every sector, every aspect of economies and societies remains aligned with human goals and values.

In this Article, we first consider the landscape of governance efforts that have been directed at AI and articulate the limitations of each approach. These limitations are twofold. The first is a technical limitation: It is deeply challenging to establish adequate controls on AI systems using the conventional tools of text-based statutes, regulations, and judicial decisions that are typically produced and enforced by politics, bureaucracy, and litigation. Indeed, this is why, as we will discuss, almost all government efforts to regulate AI to date have turned to the use of technical standard-setting organizations, outside of government. But this introduces the second limitation of current efforts: the democratic limitation. Many, if not most, of the regulatory imperatives with AI are not technocratic in nature; they are deeply value-laden choices. How do we trade off the gains of increased efficiency and innovation afforded by AI systems against risks to economic or political stability, or the health and welfare of individuals and groups? These are trade-offs that modern societies make through complex schemes of law and politics that are ultimately accountable to citizens. This is a

4. *See, e.g.*, Latanya Sweeney, *Discrimination in Online Ad Delivery*, COMMC’NS ACM, May 2013, at 44; Joy Buolamwini & Timnit Gebru, *Gender Shades: Intersectional Accuracy Disparities in Commercial Gender Classification*, 81 CONFERENCE ON FAIRNESS, ACCOUNTABILITY, AND TRANSPARENCY 1 (2018), http://proceedings.mlr.press/v81/buolamwini18a/buolamwini18a.pdf [https://perma.cc/EX52-9U7R].

5. *See generally* Anna Jobin et al., *The Global Landscape of AI Ethics Guidelines*, 1 NATURE MACH. INTEL. 389 (2019).

6. Gary E. Marchant, *Addressing the Pacing Problem*, *in* THE GROWING GAP BETWEEN EMERGING TECHNOLOGIES AND LEGAL-ETHICAL OVERSIGHT: THE PACING PROBLEM 199 (2011).

7. Indeed, our proposal for regulatory markets for AI safety was originally released as a preprint in December 2019, a few months following OpenAI’s release of the large language model GPT2. Jack Clark & Gillian K. Hadfield, *Regulatory Markets for AI Safety*, 1, 1 ARXIV (Dec. 11, 2019), https://arxiv.org/pdf/2001.00078 [https://web.archive.org/web/20250518102325/https://arxiv.org/pdf/2001.00078]. Since then, there have been several major further generations of language models (GPT5/Claude Opus 4/Gemini 2.5) as well as the emergence of entirely new AI capabilities like widely available text-to-video and agentic systems.

major challenge for AI governance in democratic market-based economies, where governments have limited influence over industry, a problem not faced by state-capitalist economies like China's.[8]

In response to this challenge, we propose a new approach to regulation to enrich the AI governance landscape. This approach attempts to simultaneously address the technical deficit of conventional regulation and the democratic deficit of industry standards. We call this approach *regulatory markets*, building on a model introduced by Hadfield to address the challenges of the modern global economy.[9] This model falls between the polar extremes of regulation entirely by technically constrained governments and entirely by democratically unconstrained private actors. As we will explain, this approach builds on new governance approaches that advocate for a shift from prescriptive, command-and-control, regulation to outcomes-based regulation, and goes further into the territory of effective regulation than other new governance approaches such as enforced self-regulation and risk-based or management-based regulation.[10]

In short, the model proposes the development of an *independent sector* of *licensed private regulators*. In this model, governments set the required outcomes for regulation; these could be metrics based (e.g., frequency of fraudulent transactions approved or of illegal content online) or principles based (e.g., reasonably low incidence of accidents in autonomous vehicles or prohibitions on hiring practices that have an unjustifiable disparate impact on protected groups). Governments license and audit private regulators, evaluating their regulatory services against the required outcomes—can the private regulator seeking a license to regulate the use of AI in hiring, for example, demonstrate achievement of the government-defined objective of nondiscrimination or demographic representation? The targets of regulation (social media platforms, banks, autonomous vehicle manufacturers or operators, employers, etc.) are required by governments to purchase the regulatory services of appropriate licensed private regulators. Governments establish rules governing private regulators to promote competition between them and uphold the integrity of regulation, ensuring that private regulators achieve the goals set by government.

Private regulators could use conventional regulatory tools—establishing written requirements, monitoring for compliance with those requirements, and penalizing violations—but we also expect that private regulators would develop new *regulatory technologies*.[11] Such technologies could themselves be data-

---

8. *See* ROBERT D. ATKINSON ET AL., CHINESE STATE CAPITALISM: DIAGNOSIS AND PROGNOSIS (Scott Kennedy & Jude Blanchette eds., 2021).

9. GILLIAN K. HADFIELD, RULES FOR A FLAT WORLD: WHY HUMANS INVENTED LAW AND HOW TO REINVENT IT FOR A COMPLEX GLOBAL ECONOMY ch. 10 (2017).

10. *See* JOHN BRAITHWAITE & PETER DRAHOS, GLOBAL BUSINESS REGULATION (2000); Christopher Carrigan & Cary Coglianese, *The Politics of Regulation: From New Institutionalism to New Governance*, 14 ANN. REV. POL. SCI. 107 (2011).

11. *See* HADFIELD, *supra* note 9. Joskow used the idea of regulatory technology fifty years ago to mean the "organizational structures, standard operating procedures, and legal rules" used in public utility regulation. Paul L. Joskow, *Inflation and Environmental Concern: Structural Change in the Process of Public Utility Price Regulation*, 17 J.L. & ECON. 291, 323 (1974). The idea of intensive and deploy AI methods. For example, a private regulator of banking services could build machine-learning-based tools and require the banks it regulates to enable regular automated sampling and auditing of the bank's transactions and the performance of the bank's AI systems. Competition between private regulators, we believe, can help to find the right balance between promoting innovation in AI technology and ensuring new technology continues to be accountable and aligned with democratically established goals and values. Driving investment and innovation in regulatory technologies is a key reason to add regulatory markets to the governance of AI toolkit. We think such technologies are essential and they are unlikely to be built in the public sector.

An additional motivation for this proposal is the challenge of developing regulation that can operate at global scale. Current efforts to establish global standards founder on the difficulty of aligning diverse nations on shared regulatory outcomes beyond very high-level abstract principles.[12] Or they fall prey to the democratic deficit of leaving the development of more specific standards to international industry organizations.[13] Our proposal aims to provide another way to thread the global needle. We expect that private regulators could operate at global scale, obtaining licenses from the political jurisdictions (nation, province or state, city) in which they seek to operate. Private regulators could diversify and tailor their offerings to divergent outcomes requirements. Governments could choose when to align their licensing requirements with those of other governments (to obtain private regulatory services in their local markets) and when to set different requirements, perhaps at the cost of regulating from within government or using private regulatory services that are more costly because of departures from dominant approaches. Private regulators also need not coordinate their efforts globally. Governments and regulators can craft standards and develop "mutual recognition" regimes, which allow one jurisdiction to recognize a common standard tested by a private regulator in another jurisdiction.

We do not think that regulatory markets will work in all circumstances. Nor will they be an appropriate approach in all circumstances. But we do think that governments must act swiftly to bolster their AI governance options by including a role for regulatory markets. And we are confident that governments and industry are ready to take this next step. In a 2024 *Wall Street Journal* essay,

---

regulatory technology extending to the use of machines such as automated data processing and AI has been most extensively developed in the context of financial regulation. For early contributions, see Douglas W. Arner et al., *FinTech, RegTech and the Reconceptualization of Financial Regulation*, 37 NW J. INT'L L. & BUS. 371 (2017); SANJAY PODDER ET AL., REGTECH FOR REGULATORS: REARCHITECT THE SYSTEM FOR BETTER REGULATION (2018).

12. Alan O. Sykes, *The (Limited) Role of Regulatory Harmonization in International Goods and Services Markets*, 2 J. INT'L ECON. L. 49 (1999).

13. *See* Michael Veale & Frederik Zuiderveen Borgesius, *Demystifying the Draft EU Artificial Intelligence Act—Analysing the Good, the Bad, and the Unclear Elements of the Proposed Approach*, 22 COMPUT. L. REV. INT'L 97, 105 (2021); *cf.* Waheed Hussain & Jeffrey Moriarty, *Accountable to Whom? Rethinking the Role of Corporations in Political CSR*, 149 J. BUS. ETHICS 519, 531 (2018).

former Google CEO Eric Schmidt—long a proponent of leaving AI regulation to industry alone—advocated for a regulatory market in AI testing companies.[14]

The Article is organized as follows. Part I gives a brief overview of the reasons why governance of AI is necessary. This includes the risks of harm that have been widely discussed in the AI governance arena, such as bias and polarization, but it also emphasizes the often-overlooked disruptions that AI could produce in a wide variety of existing regulatory schemes such as oversight of health care, medical devices, and pharmaceuticals. In Part II, we give a brief, overview of the state of AI governance as of 2025, reviewing the initiatives and models that have emerged around the world. Part III introduces the model of regulatory markets in more detail, provides some examples of how it could function in practice, and considers what is required from governments to establish and regulate these markets. In Part IV, we discuss limitations and risks of the model and make the case for why the model offers a needed expansion of the AI governance approaches currently in play around the globe. Following the main Article's focus on Western AI governance regimes, Part IV also addresses the important question of how regulatory markets might serve as a model that can support global trade rules that generalizes to China. Part V provides concluding remarks.

## I. THE AI GOVERNANCE CHALLENGE

There is little consensus about a detailed definition of artificial intelligence. Some use the term *AI* only to refer to the machine-learning systems that have played a prominent role in producing the significant advances of the past decade;[15] others emphasize that AI is a suite of computational techniques that includes the classical rules-based and symbolic systems that predated the explosion of deep learning.[16]

For the purposes of policy, it is less important to be specific about computational techniques than it is to focus on which kinds of AI systems produce new challenges for governance. Here, a key consideration is the extent to which a computational system functions reliably in ways that were contemplated and intended by its designer. This is how purely mechanical systems function: An internal-combustion engine is designed to use a spark to ignite compressed fuel that changes pressure in a cylinder that moves a piston that moves gears and then wheels. Ensuring such systems function safely and as intended is something we have learned to accomplish fairly well through testing and, when such systems are operated by complex human-coded software, through formal verification techniques.[17]

---

14. Eric Schmidt, *How We Can Control AI*, WALL ST. J. (Jan. 26, 2024, 11:06 AM), https://www.wsj.com/tech/ai/how-we-can-control-ai-327eeecf [https://perma.cc/Y2NW-EHQ5].

15. STUART RUSSELL & PETER NORVIG, ARTIFICIAL INTELLIGENCE: A MODERN APPROACH (4th ed., 2021).

16. NILS J. NILSSON, THE QUEST FOR ARTIFICIAL INTELLIGENCE (2010).

17. *See, e.g.*, Dolores R. Wallace & Roger U. Fujii, *Software Verification and Validation: An Overview*, IEEE SOFTWARE, May 1989, at 10.

But as we build more intelligent systems, the distance between what a human system designer contemplated and intended increases, and the behavior of the system becomes harder to predict, test and channel. There are two essential reasons. One, AI systems are valuable precisely because they are (ideally) capable of solving problems that humans cannot solve—processing massive quantities of data at rates much faster than humans can achieve; discerning patterns that humans do not see. The goal in many cases is superhuman performance.[18] If a machine displays intelligence that surpasses that of its designers, it can be very challenging to know how to predict and check it or how to intervene to modify it. This is not to say it is impossible, just that it is a substantial, and substantially new, governance challenge. This is also why useful definitions of AI from a governance point of view, like the OECD's, emphasize the *quasi-autonomy* of these systems: Even if they merely recommend actions to humans, humans who cannot match the system's reasoning capabilities may change their own reasoning and behavior, and hence the environment, as a result of receiving the recommendation.[19]

The second reason the distance between designer intent and system behavior grows with AI is that intelligent systems can be general in application; they can be deployed in environments and put to uses not contemplated by their designers. The engineers who develop a facial recognition or large language model cannot possibly anticipate and test for all the different ways in which the models could be used. Indeed, the great promise of AI lies in its generality. Even if it is very long before we see systems that might go under the name of artificial general intelligence (sometimes defined as able to do any cognitive task a human can), AI is already understood as a *general-purpose technology*,[20] a *method of inventing inventions*.[21] Moreover, increasingly, the barrier to entry for using this technology to invent something new is quite low. This low barrier to entry means designers of AI are putting tools on the market that they themselves cannot possibly test and validate in all uses; and users that did not design the tools may be too far from the design and conception phase to contemplate all the ways in which they could go astray in a particular context. Moreover, controls to shape the behavior of the technology in the wide expanse of possible applications can be hard to design and reliably implement, by either developer or deployer.

---

18. Adam Clark, *Meta Is Betting Its Future on 'Personal Superintelligence.' Here's What That Means.*, BARRON'S (July 31, 2025, 11:14 AM), https://www.barrons.com/articles/meta-platforms-stock-ai-personal-superintelligence-ab08bc6a?utm_source=chatgpt.com.

19. Organisation for Economic Co-operation and Development, *Scoping the OECD AI Principles: Deliberations of the Expert Group on Artificial Intelligence at the OECD (AIGO)* (OECD Digital Economy Papers No. 291, 2019).

20. Erik Brynjolfsson et al., *Artificial Intelligence and the Modern Productivity Paradox: A Clash of Expectations and Statistics*, *in* THE ECONOMICS OF ARTIFICIAL INTELLIGENCE: AN AGENDA 23 (Ajay Agrawal et al. eds., 2018) [hereinafter ECONOMICS OF AI].

21. Iain M. Cockburn et al., *The Impact of Artificial Intelligence on Innovation: An Exploratory Analysis*, *in* ECONOMICS OF AI, *supra* note 20, at 115 (2018).

For both reasons, the machine-learning models that have dominated the headlines of the past decade are a good paradigm for the AI governance challenge, even if they are not the only form of advanced AI we can expect. We will focus on this paradigm through the rest of the Article.

Machine-learning models epitomize the disconnect between human intent and system behavior because the software that is ultimately deployed (to recognize a face, to summarize a text, to recommend a medical diagnosis) is written not by a human but by the machine itself.[22] Machine-learning models are given massive quantities of data and human-designed algorithms that instruct the machine how to use the data to build a *model* that accomplishes a statistical objective. Such objectives may include (1) classifying objects to match the human-generated labels on a training set; (2) exploring an environment to discover which actions maximize a human-designated reward function; (3) clustering objects in groups to minimize the extent to which each member of a group differs from the others; (4) reliably predicting from some elements of a collection of words or an image what the remaining elements probably are, and (5) inferring from human behavior what the humans probably value. The resulting software (which takes in new data, conducts the mathematical operations contained in the model, and produces output) is the end product of that process. The model may be simple to understand: If we provide the machine with historical data structured with human-defined variables—age, gender, credit score, for example—and instruct the machine to perform a linear regression to predict another variable—probability of repaying a loan, for example—we can easily understand how the resulting four-parameter model (three coefficients and a constant) will

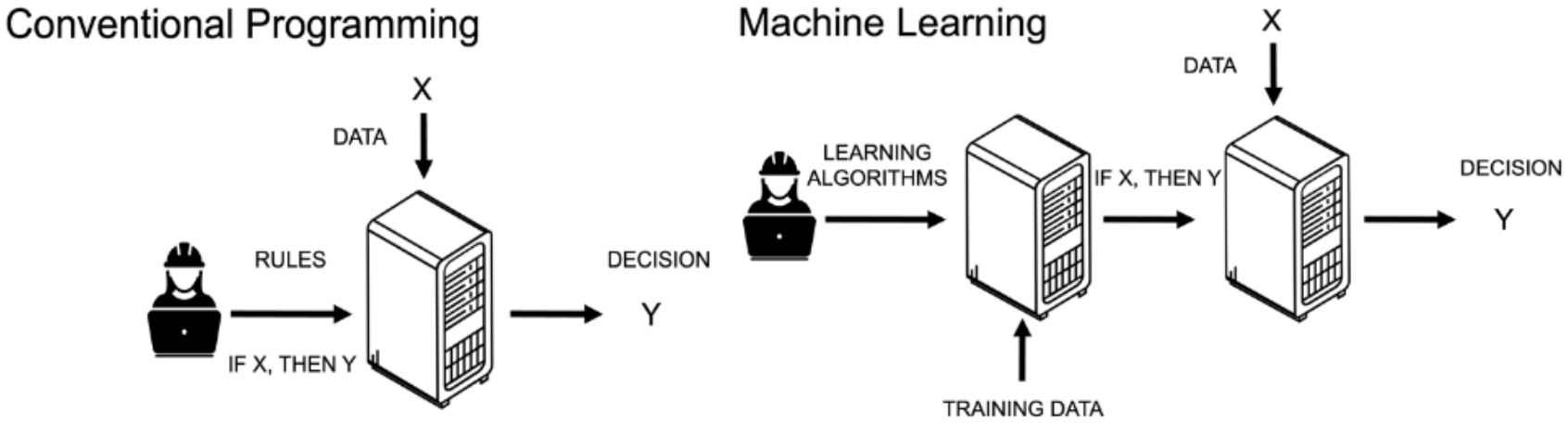

**Figure 1. Conventional Programming and Machine Learning.** Conventional programming uses human-coded rules to transform input data into decisions or outputs; machine learning uses human-coded learning algorithms that process training data to produce machine-generated rules to transform input data into decisions or outputs.

behave, and what pitfalls it might contain, if we then use it to decide to whom to give credit. We can also easily intervene to modify the predictions of the model if there are reasons to believe that the data has captured effects that we want, as a matter of policy, to override or modify. But as a general matter, and

22. *See infra* Figure 1.

increasingly, machine-learning models are complex and difficult to interpret and intervention to modify their behavior can be very difficult. As noted earlier, some of the most striking results of the past few years have come from large language models with billions or even trillions of parameters.[23]

It is the gap between what a human designer intends and the way complex AI systems can behave that generates the AI governance challenge. In many domains of regulation, we can regulate the designers, that is the programmers, to ensure that they are taking reasonable steps to produce a product that behaves as society demands—so not unlawfully discriminating between groups, not causing people to suffer physical harm, and not misleading consumers. We have lots of experience to draw on to determine what it is reasonable to ask producers to do, and we have lots of science and experimental testing methods and evaluation criteria to draw on. Regulation of pharmaceuticals, for example, draws on centuries of methodology and statistics to reliably predict how drugs will behave in the body. Regulation of automobiles similarly draws on extensive experience with harms from auto accidents and extensive engineering science and expertise to reliably predict how a design modification will impact the likelihood of injuries. But predicting, evaluating, and constraining the behavior of an AI system requires experience we have not yet gained and novel methods that largely do not yet exist. This is not to say that we cannot identify harms or behaviors or outcomes we want to avoid or prevent entirely. Nor is it to say that there is nothing we can currently do or legal rules that we can apply.[24] But fundamental gaps in predictability and knowledge, at least in the current state of the scientific and experiential record, imply that regulatory methods to implement those expectations are themselves in need of extensive research and development. As shown in the next Part, governments are struggling to meet the AI governance challenge effectively.

## II. THE AI GOVERNANCE LANDSCAPE CIRCA 2025

For most of the past decade, discussion of AI governance in Western democracies has focused on what Gutierrez and Marchant call "soft law."[25] This has been the era of "AI ethics."[26] This approach to governance looks to broad principles, ethical mandates, and voluntary codes of conduct to steer AI development and deployment in desired directions. As of 2019, Gutierrez and

---

23. *See, e.g.*, Deep Ganguli et al., *Predictability and Surprise in Large Generative Models*, *in* 2022 ACM CONFERENCE ON FAIRNESS, ACCOUNTABILITY, AND TRANSPARENCY 1747 (2022).

24. *See, e.g.*, Joanna J. Bryson, *The Artificial Intelligence of the Ethics of Artificial Intelligence: An Introductory Overview for Law and Regulation*, *in* THE OXFORD HANDBOOK OF ETHICS OF AI 2 (2020).

25. CARLOS IGNACIO GUTIERREZ & GARY E. MARCHANT, A GLOBAL PERSPECTIVE OF SOFT LAW PROGRAMS FOR THE GOVERNANCE OF ARTIFICIAL INTELLIGENCE 3 (2021), https://lsi.asulaw.org/softlaw/wp-content/uploads/sites/7/2022/08/final-database-report-002-compressed.pdf [https://web.archive.org/web/20250602091429/https://lsi.asulaw.org/softlaw/wp-content/uploads/sites/7/2022/08/final-database-report-002-compressed.pdf].

26. *See* Julia Black & Andrew Murray, *Regulating AI and Machine Learning: Setting the Regulatory Agenda*, 10 EUR. J.L. & TECH., no. 3, 2019, at 1, https://www.ejlt.org/index.php/ejlt/article/view/722/980 [https://perma.cc/9G5W-737M].

Marchant identified 634 soft law programs[27] (defined as programs that set out "substantive expectations that are not directly enforceable by government[s]"[28]), with approximately ninety-five percent published between 2015 and 2019.[29] AlgorithmWatch created a global inventory of AI Ethics Guidelines, which contained 167 frameworks and principles as of April 2020.[30] Frameworks and principles have been produced by governments, civil society organizations, corporations, industry organizations, and professional associations.[31] Most are fairly abstract and there is substantial overlap in content, with convergence around five high-level concepts: transparency, justice and fairness (non-bias, nondiscrimination), non-maleficence (security, safety, non-subversion), responsibility and accountability, and privacy.[32]

Gutierrez and Marchant predicted in 2021 that soft law "will be the dominant form of artificial intelligence (AI) governance for the foreseeable future."[33] And as late as 2019, leaders in the tech industry were still taking the position that "self- and co-regulatory approaches . . . have been largely successful at curbing inopportune AI use."[34] But highly publicized problems ranging from the Cambridge Analytica scandal, in which millions of Facebook profiles were harvested and used for political influence in the 2016 U.S. presidential election, to the abuse of AI in policing and immigration[35] triggered calls for a shift from broad-brush voluntary ethical codes and principles to more formal regulatory tools beginning around 2020. Then the dramatic introduction of ChatGPT in late 2022, exposing millions around the world to the potential, and risks, of powerful large language models, increased the sense of urgency around regulation. Governments responded, broadly speaking, in two ways: either by enacting AI-specific legislation that cuts across existing regulatory domains or by formalizing reliance on broad principles and voluntary industry-led standards, leaving specific legislation to existing sectoral regulators. Almost across the board, however, whether mandated or voluntary, the approach to AI regulation, at least in the West and thus far, falls into the category of *management-based regulation*

27. GUTIERREZ & MARCHANT, *supra* note 25, at 3.

28. *Id.* at 5.

29. *Id.* at 9 tbl.3.

30. *AI Ethics Guidelines Global Inventory*, ALGORITHMWATCH, https://inventory.algorithmwatch.org/ [https://perma.cc/RH4C-3MW8].

31. *About AI Ethics Guidelines Global Inventory*, ALGORITHMWATCH, https://inventory.algorithmwatch.org/about [https://perma.cc/4PTA-SVGX].

32. Jobin et al., *supra* note 5.

33. GUTIERREZ & MARCHANT, *supra* note 25, at 3.

34. GOOGLE, PERSPECTIVES ON ISSUES IN AI GOVERNANCE 2 (2019), https://ai.google/static/documents/perspectives-on-issues-in-ai-governance.pdf [https://perma.cc/WGU3-9KAS].

35. *See* MEREDITH WHITTAKER ET AL., AI NOW INST., AI NOW REPORT 2018 (2018), https://ainowinstitute.org/wp-content/uploads/2023/04/AI_Now_2018_Report.pdf [https://perma.cc/Z656-H34E].

(also called process-oriented, risk-based, or enforced self-regulation), which requires (or encourages) firms to evaluate the risks generated by their business and to develop their plan for how those risks will be managed.[36]

## A. The E.U. Approach: Comprehensive AI Legislation

The European Union has led the first, comprehensive approach to AI legislation. These efforts began with arguably the first AI legislation seen globally, in the form of a provision in the European Union's 2018 General Data Protection Regulation (GDPR).[37] The GDPR requires entities that use automated decision systems to disclose this fact;[38] to provide "meaningful information about the logic involved"[39] and a right to obtain human intervention and to contest the decision;[40] and not to use "special categories of personal data" such as race and gender unless an exceptional lawful basis for processing such data is present.[41] Subsequently, in 2022, the European Union passed two key pieces of legislation. The first is the Digital Services Act, which regulates very large online platforms and the AI systems they use to recommend content and deliver search results (requiring, for example, that platforms identify the most important parameters used to rank content for a particular user).[42] The second is the Digital Markets Act, which regulates the AI-powered "gatekeeping" providers of core platform services in the digital economy to ensure that they do not distort market competition (requiring, for example, that automated ranking of consumer search results not prioritize the gatekeeper's own products and that the gatekeeper's access to aggregate data not be used to compete unfairly with business users of its platform).[43]

---

36. *See* Sharon Gilad, *Process-Oriented Regulation: Conceptualization and Assessment*, *in* HANDBOOK ON THE POLITICS OF REGULATION 423 (2011); John Braithwaite, *The Essence of Responsive Regulation*, 44 U.B.C. L. REV. 475 (2011) [hereinafter Braithwaite, *The Essence of Responsive Regulation*]; Cary Coglianese et al., *Performance-Based Regulation: Prospects and Limitations in Health, Safety, and Environmental Protection*, 55 ADMIN. L. REV. 705 (2003); IAN AYRES & JOHN BRAITHWAITE, RESPONSIVE REGULATION: TRANSCENDING THE DEREGULATION DEBATE (1992); John Braithwaite, *Enforced Self-Regulation: A New Strategy for Corporate Crime Control*, 80 MICH L. REV. 1466 (1981).

37. General Data Protection Regulation, Council Regulation No. 2016/679, 2016 OJ (L 119) 1.

38. *Id.* art. 22.

39. *Id.* art. 13.2(f), art. 14(2)(g), art. 15(1)(h).

40. *Id.* art. 22(3).

41. *Id.* art. 22(4).

42. Regulation (EU) 2022/2065 of the European Parliament and of the Council of 19 October 2022 on a Single Market for Digital Services and amending Directive 2000/31/EC (Digital Services Act), 2022 O.J. (L 277) 1 [hereinafter Digital Services Act].

43. Regulation (EU) 2022/1925 of the European Parliament and of the Council of 14 September 2022 on Contestable and Fair Markets in the Digital Sector and Amending Directives (EU) 2019/1937 and (EU) 2020/1828 (Digital Markets Act), 2022 O.J. (L 265) 1 [hereinafter Digital Markets Act].

Most comprehensive, however, is the European Union's AI Act, proposed in 2021[44] and finally enacted in 2024.[45] This is sweeping legislation that is directly focused on the regulation of AI systems. The E.U. AI Act applies to any AI system placed on the market or put into service in the European Union, regardless of where providers are physically located.[46] Penalties for violation of the E.U. AI Act can be for as much as six percent of global annual revenues.[47]

The scheme of the E.U. AI Act is risk based. For application-specific AI systems it differentiates between unacceptable, high, and low-risk uses. "Risk" is defined to mean "the combination of [both] the probability . . . and the severity of [] harm."[48] Unacceptable risk uses (subliminal manipulation, exploitation of vulnerable groups, social scoring, some real-time uses of biometrics by law enforcement) are prohibited.[49] High-risk uses include products that are already regulated by the European Union such as machinery, toys, medical devices, and transportation vehicles and the safety components of critical infrastructure.[50] A product or service impacting education, employment, public benefits, credit, law enforcement, border control, or judicial or democratic processes is deemed high risk unless "it does not pose a significant risk of harm to the health, safety or fundamental rights of natural persons, including by not materially influencing the outcome of decision making."[51]

General-purpose AI models are deemed to have "high-impact capabilities" if their capabilities "match or exceed the capabilities recorded in the most advanced general purpose AI models,"[52] a threshold that is presumed if the computation used to train the model exceeds $10^{25}$ floating point operations (FLOPs).[53] General-purpose models with high impact capabilities are classified as models with "systemic risk" meaning they pose a risk of "significant impact on the [European] Union due to their reach or due to actual or reasonably foreseeable negative effects on public health, safety, public security, fundamental rights, or the society as a whole, that can be propagated at scale across the value chain."[54]

All providers of AI systems—high- or low-risk, narrow- or general-purpose—must ensure that people know they are interacting with an AI system if it would

44. *Proposal for a Regulation of the European Parliament and of the Council Laying Down Harmonised Rules on Artificial Intelligence (Artificial Intelligence Act) and Amending Certain Union Legislative Acts EXPLANATORY MEMORANDUM*, COM/2021/206 final COM (2021) 206 final (Apr. 21, 2021) [hereinafter Proposed E.U. AI Act].

45. Regulation (EU) 2024/1689 of the European Parliament and of the Council of 13 June 2024 Laying Down Harmonised Rules on Artificial Intelligence and Amending Regulations (EC No 300/2008, (EU) No 167/2013, (EU) No 168/2013, (EU) 2018/858, (EU) 2018/1139 and (EU) 2019/2144 and Directives 2014/90/EU, (EU) 2016/797 and (EU) 2020/1828, 2024 O.J. (L) 1 [hereinafter E.U. AI Act].

46. *Id.* art. 2.

47. Proposed E.U. AI Act, *supra* note 44, art. 71(3).

48. E.U. AI Act, *supra* note 45, art. 3(2).

49. *Id.* art. 5.

50. *Id.* annex I, II.

51. *Id.* art. 6(3).

52. *Id.* art. 3(64).

53. *Id.* art. 51(2).

54. *Id.* art. 3(63–65).

not be obvious to a reasonable user and must disclose the presence of synthetic audio, image, video or text content and emotion-recognition and biometric categorization systems.[55] Beyond this, low-risk systems are encouraged but not required to adopt codes of conduct based on the E.U. AI Act.[56]

High-risk AI systems, including systems in listed high-risk domains (such as education and employment) which providers have concluded do not pose a high risk because they do not pose a significant risk of harm to health, safety or fundamental rights, must be registered on a public E.U. database.[57] High-risk systems are required to have in place a risk-management system and practices to ensure data is of adequate quality and that systems are adequately documented, accurate, robust and secure against intrusion and manipulation.[58] They must be sufficiently transparent so that users can understand the system and use it appropriately, and there must be post-market monitoring.[59] High-risk systems must also have in place practices that enable traceability (recordkeeping), human oversight, and a quality management system to ensure compliance with E.U. AI Act requirements.[60] Most providers of high-risk systems can self-certify their compliance with the E.U. AI Act (and are required to mark the system documentation with a symbol to this effect); but providers of AI systems that contribute to critical infrastructure must have conformity confirmed by a third-party certifying body approved by a member state.[61]

Unless they make their models available under a "free and open-source licence," providers of general-purpose AI models are required to maintain up-to-date technical documentation of the training and evaluation of the model to be provided to E.U. authorities on request.[62] AI system providers must also give users information, which provides a "good understanding of the capabilities and limitations of the general-purpose AI model"[63] and ensure their own compliance with the E.U. AI Act. In addition, and regardless of whether the model is closed- or open-source, providers of general-purpose models with systemic risk are required to notify the European Commission that they have met the definition of systemic risk, perform state-of-the-art evaluations of model capabilities and conduct adversarial testing (red teaming) of models, assess and mitigate possible systemic risks, track and where possible correct serious incidents, and ensure adequate levels of cybersecurity.[64]

A critical feature of the E.U. approach is the anticipated role for private standard-setting bodies. The International Organization for Standardization (ISO) and the Institute of Electrical and Electronics Engineers (IEEE) both had begun to publish AI-specific standards as of 2021.[65] Guttierez and Marchant include standards in their catalogue of soft law programs; as of 2019, ten percent of the programs they identified were classified as standards.[66] The E.U. AI Act allows the European Commission to ask private European standardization bodies to create harmonized standards to provide technical implementation of the Act's requirements; compliance with such standards will then be deemed to be presumptive compliance with the E.U. AI Act. The standard-setting process was initiated in May 2023 and is projected for completion in 2026.[67] Until harmonized standards are published, industry can show presumptive compliance with the Act by following an applicable code of practice.[68] A code of practice for general-purpose AI models was published in 2025; initial signatories to these voluntary requirements included Anthropic, OpenAI, Google, and Microsoft.[69] The code was developed through a multi-month consultation led by the E.U. AI Office, which included companies, industry groups, academics, and representatives of civil society.[70] A crucial aspect of the code is commitments by the companies to publish more details about how they test and evaluate their AI systems for various properties relative to public safety, as well as details on how and who they work with for third-party evaluation or analysis of compliance.[71]

---

55. *Id.* art. 50.
56. *Id.* art. 95.
57. *Id.* art. 49.
58. *Id.* arts. 9, 12, 13, 15.
59. *Id.* arts. 13, 72.
60. *Id.* arts. 12, 14, 17.
61. Proposed E.U. AI Act, *supra* note 44, art. 43. Products that are otherwise regulated by the E.U. such as machinery and medical devices must also obtain third-party conformity assessments if required by the relevant product-specific legislation.
62. E.U. AI Act, *supra* note 45, art. 53.
63. *Id.* art. 53(1)(b).
64. *Id.* art. 55.

### B. The U.S./U.K. Approach: Existing Regulators and Voluntary Standards

In contrast to the European Union, the United States and the United Kingdom have chosen, thus far, to look primarily to existing regulators to develop regulations pertaining to AI in their particular domain (transportation, finance, consumer protection, etc.) and to otherwise rely on, and participate in the development of, voluntary international standards in bodies such as the ISO and the IEEE. Again, the release of ChatGPT in late 2022 changed the direction somewhat, with both countries beginning to take more concrete steps to address systemic risks generated by frontier models.

---

65. *See* GUTIERREZ & MARCHANT, *supra* note 25.

66. *Id.* at 11, tbl. 6. Approximately half of the standards Gutierrez and Marchant identify pertained to autonomous vehicles.

67. JOSEP SOLER GARRIDO ET AL., JRC 139430, HARMONISED STANDARDS FOR THE EUROPEAN AI ACT 1, 2 (2024), https://publications.jrc.ec.europa.eu/repository/handle/JRC139430 (click "Download" to access policy brief).

68. E.U. AI Act, *supra* note 45, art. 56(4).

69. *The General-Purpose AI Code of Practice*, EUR. COMM'N, https://digital-strategy.ec.europa.eu/en/policies/contents-code-gpai [https://perma.cc/D67J-SN38].

70. *Meet the Chairs Leading the Development of the First General-Purpose AI Code of Practice*, EUR. COMM'N (Sept. 30, 2024), https://digital-strategy.ec.europa.eu/en/news/meet-chairs-leading-development-first-general-purpose-ai-code-practice [https://perma.cc/7EQG-JVBP]; *Drawing-up a General-Purpose AI Code of Practice*, EUR. COMM'N, https://digital-strategy.ec.europa.eu/en/policies/ai-code-practice [https://perma.cc/Y4YJ-R4YE].

71. *See The General-Purpose AI Code of Practice*, *supra* note 69.

*United States:* In the United States, which is by far the dominant jurisdiction for AI research and development,[72] the pre-2023 policy was first expressed in a presidential executive order (E.O.), issued in 2019. This E.O. emphasized the importance of supporting "continued American leadership in AI."[73] The E.O. directed the Office of Management and Budget (OMB) to create guidance for regulation of AI applications to inform the development of regulatory approaches "that advance American innovation while upholding civil liberties, privacy and American values" and the National Institute of Standards and Technology (NIST) to "issue a plan for Federal engagement in the development of technical standards."[74]

The OMB guidance issued in 2020 directed regulatory agencies to "avoid regulatory or non-regulatory actions that needlessly hamper AI innovation and growth."[75] It then set out principles to guide regulation, including a focus on public trust through the promotion of reliable, robust, and trustworthy AI and protection of reasonable expectations of privacy, nondiscrimination, safety and security, and transparency to enable understanding of how an AI application or system works. The OMB guidance called for the use of risk assessment and management approaches that avoid "unnecessarily precautionary approaches to regulation that unjustifiably create anticompetitive effects or inhibit innovation"[76] and encouraged flexible, performance-based approaches "that are technology neutral and that do not impose mandates on companies that would harm innovation,"[77] including "[t]argeted agency conformity assessment schemes, to protect health, safety, privacy and other values."[78]

NIST issued a plan in 2019 for U.S. participation in developing technical standards, which called for research into standards development and the development of "metrics and data sets to assess reliability, robustness and other trustworthy attributes of AI systems"[79] that could be incorporated into standards, support for public-private partnerships to develop innovative approaches to standards, and participation in international standard-setting efforts "to advance

72. INST. FOR HUMAN-CENTERED A.I., STANFORD UNIV., ARTIFICIAL INTELLIGENCE INDEX REPORT (2024), https://hai-production.s3.amazonaws.com/files/hai_ai-index-report-2024-smaller2.pdf [https://perma.cc/H4NT-XY9F].

73. Maintaining American Leadership in Artificial Intelligence, Exec. Order No. 13859, 84 Fed. Reg. 3967, 3967 (2019).

74. *Id.* at 3970.

75. Memorandum from Russell T. Vought, Dir., Off. of Mgmt. & Budget, on Guidance for Regulation of Artificial Intelligence Applications 2 (Nov. 17, 2020), https://www.whitehouse.gov/wp-content/uploads/2020/11/M-21-06.pdf [https://perma.cc/BE69-LZ7E].

76. *Id.* at 4–5.

77. *Id.* at 5.

78. *Id.*

79. NAT'L INST. OF STANDARDS & TECH., U.S. LEADERSHIP IN AI: A PLAN FOR FEDERAL ENGAGEMENT IN DEVELOPING TECHNICAL STANDARDS AND RELATED TOOLS 5 (2019), https://www.nist.gov/system/files/documents/2019/08/10/ai_standards_fedengagement_plan_9aug2019.pdf [https://perma.cc/9UHJ-PVCN].

AI standards for U.S. economic and security needs."[80] NIST's AI Risk Management Framework, published in January 2023,[81] was organized around familiar principles seen previously in the principles statements that began appearing in 2015:[82] validity and reliability, safety, fairness and nondiscrimination, security, resilience, accountability and transparency, explainability and interpretability, and privacy. The framework is voluntary and intended to aid organizations to manage "system risks or broader enterprise risks"[83] and "manage the design, development, deployment, evaluation, and use of AI systems."[84] It is "law- and regulation-agnostic" rather than intended as a compliance mechanism.[85]

After ChatGPT's release in late 2022, the United States significantly ramped up these administrative and voluntary efforts with the release of E.O. 14110 in October 2023.[86] E.O. 14110 required all federal agencies to appoint a Chief AI Officer to coordinate AI policy and instructed most major departments to develop steps to address the impact of AI on specific sectors or domains such as labor, education, antitrust, consumer protection, and civil rights.[87] The most expansive new steps were taken with respect to national security risks from frontier models (termed *dual-use foundation models* in the executive order).[88] Developers of these models were required to notify the federal government if they were training models with compute levels greater than $10^{26}$ FLOPs ($10^{23}$ if training on biological sequence data) and to share details about these models, such as performance on red-teaming tests to evaluate capacity to lower barriers to entry to the production of bioweapons or to engage in self-replication. Operators of large compute clusters (greater than $10^{20}$ FLOP/sec in a single data center) were required to report on their existence and location.[89] Regulations were to be developed under E.O. 14110 for reporting on foreign persons training large AI models on U.S. infrastructure and verifying the identity of foreign buyers.[90] The executive order called for NIST to expand its risk-management framework to address generative AI and dual-use foundation models[91] and amplified the call for the United States to lead on the development of international standards.

As directed by E.O. 14110, NIST released initial drafts for public comment in April 2024 on a set of voluntary standards-related initiatives: (1) a proposal

---

80. *Id.* at 5–6 (emphasis omitted).

81. Nat'l Inst. of Standards & Tech., Artificial Intelligence Risk Management Framework (AI RMF 1.0) (2023), https://nvlpubs.nist.gov/nistpubs/ai/nist.ai.100-1.pdf [https://perma.cc/AU5Q-KYX4].

82. *See* Jobin et al., *supra* note 5; Gutierrez & Marchant, *supra* note 25.

83. Nat'l Inst. of Standards & Tech., *supra* note 81, at 8.

84. *Id.* at 9.

85. *Id.* app. D at 42.

86. Exec. Order No. 14110, 88 Fed. Reg. 75191 (Oct. 30, 2023).

87. *See generally id.*

88. *Id.* at 75194.

89. *Id.* at 75197–98.

90. *See id.* at 75198–99.

91. *Id.* at 75196. The first draft of this framework was released in April 2024. Nat'l Inst. of Standards & Tech., NIST SP 800-218A, Secure Software Development Practices for Generative AI and Dual-Use Foundation Models (2024), https://nvlpubs.nist.gov/nistpubs/SpecialPublications/NIST.SP.800-218A.ipd.pdf [https://perma.cc/JRQ9-2HMM].

for creating technical approaches to detecting and tracking the provenance of synthetic content and preventing the use of generative AI to produce "child sexual abuse material or non-consensual intimate imagery of real individuals";[92] (2) a risk-management framework for generative AI that extends the existing NIST risk-management framework for AI;[93] (3) a description of secure software development practices for generative AI and dual-use foundation models;[94] and (4) a plan that sets out issues and priorities for U.S. engagement on the development of global AI standards, emphasizing the U.S. focus on industry consensus standards and the need for broad stakeholder participation.[95]

In January 2025, the Trump administration rescinded E.O. 14110 and directed federal agencies to review and rescind any actions taken under that prior order that were inconsistent with a policy focused on, among other things, U.S. global leadership in AI.[96] The absence of legislation in the United States may not last. As of mid-2025, although the likelihood of federal legislation was low in light of the AI Action Plan released in July 2025,[97] nearly all fifty states had introduced bills regulating aspects of different aspects of AI,[98] prompting attempts to impose a moratorium on state-level AI regulation.[99]

*United Kingdom:* The United Kingdom is following the same path as the United States, with an emphasis on promoting innovation, leaving AI-specific regulation to the remit of existing regulators. A March 2023 government white paper articulated the goal of focusing Parliament only on developing cross-cutting high-level principles to coordinate these regulatory efforts, based on the OECD principles.[100] The government said it did not currently see a need for regulation

92. NAT'L INST. OF STANDARDS & TECH., NIST AI 100-4, REDUCING RISKS POSED BY SYNTHETIC CONTENT 5 (2024), https://airc.nist.gov/docs/NIST.AI.100-4.SyntheticContent.ipd.pdf [https://perma.cc/64QH-LM8T].

93. NAT'L INST. OF STANDARDS & TECH., NIST AI 600-1, ARTIFICIAL INTELLIGENCE RISK MANAGEMENT FRAMEWORK: GENERATIVE ARTIFICIAL INTELLIGENCE PROFILE (2024), https://nvlpubs.nist.gov/nistpubs/ai/NIST.AI.600-1.pdf [https://web.archive.org/web/20240727203623/https://nvlpubs.nist.gov/nistpubs/ai/NIST.AI.600-1.pdf].

94. NAT'L INST. OF STANDARDS & TECH., *supra* note 91.

95. NAT'L INST. OF STANDARDS & TECH., NIST AI 100-5, A PLAN FOR GLOBAL ENGAGEMENT ON AI STANDARDS (2024), https://nvlpubs.nist.gov/nistpubs/ai/NIST.AI.100-5.pdf [https://web.archive.org/web/20240727004002/https://nvlpubs.nist.gov/nistpubs/ai/NIST.AI.100-5.pdf].

96. Exec. Order No. 14148, 90 Fed. Reg. 8237 (Jan. 28, 2025); Exec. Order No. 14179, 90 Fed. Reg. 8741 (Jan. 31, 2025).

97. MICHAEL J. KRATSIOS ET AL., WINNING THE RACE: AMERICA'S AI ACTION PLAN (2025) [hereinafter WHITE HOUSE AI ACTION PLAN], https://www.whitehouse.gov/wp-content/uploads/2025/07/Americas-AI-Action-Plan.pdf [https://perma.cc/P5EH-MA2U].

98. Gregory S. Dawson et al., *How Different States Are Approaching AI*, BROOKINGS INST. (Aug. 18, 2025), https://www.brookings.edu/articles/how-different-states-are-approaching-ai [https://perma.cc/AS5V-Q57K].

99. Alasdair Phillips-Robins & Scott Singer, *The State of State AI Law: What's Coming Now that the Federal Moratorium Is Dead*, CARNEGIE ENDOWMENT (July 10, 2025), https://carnegieendowment.org/research/2025/07/state-ai-law-whats-coming-now-that-the-federal-moratorium-is-dead?lang=en.

100. *See* U.K. DEP'T FOR SCI. INNOVATION & TECH., A PRO-INNOVATION APPROACH TO AI REGULATION (2023), https://assets.publishing.service.gov.uk/government/uploads/system/uploads/attachment_data/file/1176103/a-pro-innovation-approach-to-ai-regulation-amended-web-ready.pdf

but did not close the door to future legislative action.[101] Instead it encouraged existing regulators to look to guidelines or voluntary approaches.[102] The white paper specifically declined to define AI but emphasized two key characteristics that generate regulatory challenges: the adaptiveness and autonomy of AI technologies.[103]

Like the United States, the United Kingdom is looking to global standard-setting initiatives. In 2022, the government announced the creation of an AI Standards Hub operated by the Alan Turing Institute (a nonprofit joint venture of several U.K. universities) with the participation of the British Standards Institute and the National Physical Laboratory "to lead in shaping global technical standards for Artificial Intelligence."[104] Also, like the United States, the United Kingdom did begin to take steps specifically to address risks from frontier models in late 2023 with the creation of an AI Safety Institute tasked with engaging in research and development of safety standards and tests.[105] And like the United States, the United Kingdom may move towards more cross-sectoral legislation as the power and reach of AI models grows.[106] In July 2024, the new Labour government's Prime Minister Kier Starmer said his party would "harness the power of artificial intelligence as we look to strengthen safety frameworks,"[107] and Labour has subsequently indicated a desire to put the U.K. Safety Institute on a "statutory footing."[108]

### C. The State of AI Governance in Western Democracies: Technical and Democratic Deficits

The European Union and the United States/United Kingdom have staked out different territory in the AI governance landscape. The E.U. AI Act creates

---

[https://web.archive.org/web/20250523040558/https://assets.publishing.service.gov.uk/government/uploads/system/uploads/attachment_data/file/1176103/a-pro-innovation-approach-to-ai-regulation-amended-web-ready.pdf].

101. *See id.* at 3.

102. *Id.* at 7.

103. *Id.* § 3.2.1.

104. Press Release, U.K. Dep't for Digit., Culture, Media & Sport et al., New UK Initiative to Shape Global Standards for Artificial Intelligence (Jan. 12, 2022), https://www.gov.uk/government/news/new-uk-initiative-to-shape-global-standards-for-artificial-intelligence [https://perma.cc/DX4V-2GYC].

105. *See* U.K. DEP'T FOR SCI. INNOVATION & TECH., INTRODUCING THE AI SAFETY INSTITUTE (2023), https://assets.publishing.service.gov.uk/media/65438d159e05fd0014be7bd9/introducing-ai-safety-institute-web-accessible.pdf [https://perma.cc/UTL7-BBPT].

106. *See* Ansh Bhatnagar & Devyani Gajjar, *Policy Implications of AI*, U.K. PARLIAMENT POSTNOTE 708 (Jan. 9, 2024), https://researchbriefings.files.parliament.uk/documents/POST-PN-0708/POST-PN-0708.pdf [https://web.archive.org/web/20240528172417/https://researchbriefings.files.parliament.uk/documents/POST-PN-0708/POST-PN-0708.pdf].

107. David Hughes, *King's Speech Sets Out Plan to 'Get Britain Building*,' STANDARD (July 17, 2024), https://www.standard.co.uk/news/politics/keir-starmer-government-britain-charles-prime-minister-b1171231.html [https://perma.cc/KV7A-VJG8].

108. HC Deb (26 July 2024) (752) col. 34WS (UK), https://hansard.parliament.uk/commons/2024-07-26/debates/24072618000012/AIOpportunitiesActionPlan [https://web.archive.org/web/20240917090432/https://hansard.parliament.uk/commons/2024-07-26/debates/24072618000012/AIOpportunitiesActionPlan] (text of Peter Kyle, U.K. Sec'y of State for Sci., Innovation & Tech., AI Opportunities Action Plan).

significant compliance obligations—at risk of substantial penalties (and probably at significant cost). The United States and the United Kingdom have elected, for now, to rely primarily on voluntary and sectoral approaches. But these approaches do not differ as much as might otherwise seem. With the exception of the Digital Markets Act and the Digital Services Act, all current regulatory approaches are forms of management-based or risk-based regulation, in which industry is responsible for identifying and managing the risks of their products and services. None, in our estimation, yet constitute a robust regulatory response that can meet the AI governance challenge. All display what we call *technical* and *democratic deficits.*

By a *technical deficit* we mean a lack of technical detail to inform AI developers and deployers about the required operational characteristics of the systems they build and use. This has been a persistent concern with the AI guidelines and principles that emerged over the last decade. It is one thing to say that an AI system must be "fair;" it is quite another to say what "fair" translates to in terms of system performance, especially given the multiplicity of statistical measures of fairness and the impossibility of satisfying them all.[109] Similarly, we lack operational requirements for "explainable" or "robust" AI. In conventional product safety domains, there are engineering and product quality metrics or benchmarks that are established by regulation. Canada's toy safety standards, for example, specify that "a doll, plush toy, or soft toy fails the requirements of the Toys Regulations if samples of its outer fabric, held at an angle of 45 degrees, ignite within 1 second of contact with a flame and the flame travels a distance of 127 millimetres (5 inches) in 7 seconds or less."[110] In the United States, producers of software used in aviation can demonstrate compliance with Federal Aviation Administration (FAA) safety regulations by following specific methods and tests set out in a technically detailed industry standard (DO-178C) developed and published by the private nonprofit standards association RTCA.[111]

---

109. Sorelle A. Friedler et al., *The (Im)possibility of Fairness: Different Value Systems Require Different Mechanisms for Fair Decision Making*, COMMC'NS ACM, Apr. 2021, at 136.

110. *Industry Guide to Health Canada's Safety Requirements for Children's Toys and Related Products*, GOV'T CAN., https://www.canada.ca/en/health-canada/services/consumer-product-safety/reports-publications/industry-professionals/industry-guide-safety-requirements-children-toys-related-products-summary/guidance-document.html#a331 [https://web.archive.org/web/20241204070402/https://www.canada.ca/en/health-canada/services/consumer-product-safety/reports-publications/industry-professionals/industry-guide-safety-requirements-children-toys-related-products-summary/guidance-document.html#a331] (Dec. 1, 2021).

111. *See, e.g.*, LEANNA RIERSON, DEVELOPING SAFETY-CRITICAL SOFTWARE: A PRACTICAL GUIDE FOR AVIATION SOFTWARE AND DO-178C COMPLIANCE (2013). The FAA publishes guidance that affirms that DO-178C is "an acceptable means of compliance for the software aspect of type certification [required to put an aircraft or major component into service] or TSO authorization [issued when an individual component meets minimum performance standards.]" FED. AVIATION ADMIN., FAA ADVISORY CIRCULAR NO. 20-115D, AIRBORNE SOFTWARE DEVELOPMENT ASSURANCE USING EUROCAE ED-12( ) AND RTCA DO-178( ), at 3 (July 21, 2017), https://www.faa.gov/documentLibrary/media/Advisory_Circular/AC_20-115D.pdf. [https://web.archive.org/web/20190327221831/https://www.faa.gov/documentLibrary/media/Advisory_Circular/AC_20-115D.pdf].

Whether mandatory, as in the European Union, or voluntary, as in the United States, the risk-management systems that current approaches call for do not yet establish technical performance requirements for AI systems. The E.U. AI Act and the NIST AI Risks Framework require developers to engage in the process of identifying, assessing, and mitigating risks of harm to health, safety, or fundamental rights. Providers of general-purpose AI in the European Union are also required to assess systemic risks of harm to public health, safety, public security, and the society as a whole. But these risk frameworks do not provide technical detail about what constitutes a harm to health, safety, fundamental rights, or society, much less what is an acceptable level of risk of such harm or how it is to be measured. We still do not have operational standards for what it means for AI to be unbiased, explainable, robust, or subject to human oversight.

Governments are well aware of the technical deficit of existing approaches to AI regulation. But across the globe, there is the same hope: As in the past,[112] technical details can be worked out by industry standard-setting organizations (SSOs). This is what we see in the current AI governance landscape. The most significant difference between the European approach and the British/American approach is that the European Union will require compliance with SSO standards; the United States and the United Kingdom will encourage and facilitate compliance.

But we think this expectation that SSOs will supply the technical and operational detail AI regulation requires and resolve the technical deficit is a false hope. For one, the legal frameworks governments are providing for the SSO exercise are in the domain of risk regulation. SSOs are responding with procedural solutions for a risk-management system, not technical detail about how to measure and mitigate risks. As of 2024, the ISO standard that contains requirements for AI systems (ISO/IEC 42001[113]), for example, requires that a business establish a responsible AI policy and AI objectives (such as "fairness" and "explainability"[114]); conduct impact assessments relevant to the objectives the business has established (including impacts on "human rights" and "norms, traditions, culture and values"[115]); and take steps such as ensuring competence of personnel, documentation of objectives and logging of system behavior, all relative to the policy and objectives the business has set.[116] These are important controls for responsible AI use, but they do not address the technical deficit; all of the technical detail is still left to be decided by the business internally.

---

112. *See* Emily S. Bremer, *Incorporation by Reference in an Open-Government Age*, 36 HARV. J.L. & PUB. POL'Y 131, 147–49 (2013); *see also* Nina A. Mendelson, *Private Control over Access to the Law: The Perplexing Federal Regulatory Use of Private Standards*, 112 MICH. L. REV. 737, 749 (2014) ("The Office of Management and Budget ('OMB') issued Circular No. A-119 in 1982, most recently revising it in 1998, directing agencies to rely on voluntary standards, including industry standards or consensus codes, rather than 'government-unique standards.'. . . In Section 12(d) of the National Technology Transfer and Advancement Act of 1995 (the 'NTTAA'), Congress provided that, unless inconsistent with law or impractical, all federal agencies are to use 'technical standards that are developed or adopted by voluntary consensus standards bodies . . . to carry out [the agencies'] policy objectives or activities.'").

113. INT'L ORG. FOR STANDARDIZATION, ISO/IEC 42001, INFORMATION TECHNOLOGY—ARTIFICIAL INTELLIGENCE—MANAGEMENT SYSTEM (2023).

114. *Id.* § B.5.4.

115. *Id.* § B.5.4, § B.5.5.

116. *Id.* § B.4.2, § B.5.5, § B 6.2.3, § B 6.2.8, § B.7.2.

The more fundamental problem with the expectation that SSOs will address the technical deficit in AI governance, however, is that reliance on SSOs generates what we call a *democratic deficit*. By this we mean delegating the fundamentally political tasks of reconciling important trade-offs in the design and deployment of AI systems to politically unaccountable private actors.[117] Our concern here is that governments are failing to make the hard choices about how AI should be built and deployed.

Others have already emphasized the constitutional dangers here:[118] SSOs are private entities, albeit nonprofit ones. The ISO is a nonprofit corporation operating as a global network of national standards bodies.[119] In some cases, the national standards bodies are independent of government entirely (although perhaps recognized as "the" national standards body, as with the British Standards Institute[120]); in others, they are owned or otherwise financially supported by government. (The Standards Council of Canada, for example, is a Crown corporation and is accountable to Parliament.[121]) The ISO produces standards that are developed by technical committees composed of industry experts (primarily engineers), most of whom are employed by the companies that will adopt standards.[122] Standards produced by SSOs such as ISO are proprietary: They can only be accessed by purchasing them and are not published or open for review and comment by the public.[123] Technical committees can include nonindustry stakeholders such as civil society organizations or consumer groups and the U.S. plan for global standards development, released in April 2024, emphasizes the need

117. For a similar use of the term, see Tony Porter, *The Democratic Deficit in the Institutional Arrangements for Regulating Global Finance*, 7 GLOBAL GOVERNANCE 427, 427 (2001) ("[T]he governance of global finance involves highly political conflicts that should not and cannot be resolved by technical experts or markets alone.").

118. Veale & Borgesius, *supra* note 13.

119. *Structure and Governance*, INT'L STANDARDS ORG., https://www.iso.org/structure.html [https://perma.cc/F5MK-HL6L].

120. *BSI History*, BRIT. STANDARDS INST., https://www.bsigroup.com/en-GB/about-bsi/our-history/ [https://perma.cc/S5YU-EWAL]; *BSI-British Standards Institute*, INT'L STANDARDS ORG., https://www.iso.org/member/2064.html [https://perma.cc/X6KN-GTJA].

121. *About Us*, STANDARDS COUNCIL CAN., https://scc-ccn.ca/about-us [https://perma.cc/NJL5-5GFH].

122. *Get Involved*, INT'L STANDARDS ORG., https://www.iso.org/get-involved.html [https://perma.cc/X72N-4VDV] ("One of the strengths of ISO standards is that they are created by the people that need them. Industry experts drive all aspects of the standard development process, from deciding whether a new standard is needed to defining all the technical content.").

123. For purposes of preparing this Article, for example, we purchased the fourteen AI standards that had been produced by ISO as of summer 2022, at a cost of approximately $2,500. A condition of the license purchased is that they are not copied or shared with anyone. For a detailed discussion of the democratic implications of proprietary standards, see Mendelson, *supra* note 112. A March 2024 decision of the European Court of Justice may change this status quo, at least in Europe: the Court held that standards produced by European standards bodies and referenced in European law as providing a means of demonstrating compliance with law (as is the case in the E.U. AI Act, among other product safety legislation) must be made freely available to the public. Case C-588/21 P, Public.Resources.Org and Right to Know v. Commission and Others, ECLI:EU:C:2024:201, ¶¶ 81–89 (Mar. 5, 2024), https://eur-lex.europa.eu/legal-content/EN/TXT/PDF/?uri=CELEX:62021CJ0588 [https://perma.cc/3JEC-2ADA].

for broad stakeholder engagement.[124] But as many have observed, there is substantial imbalance in the capacity for large corporate entities compared to small businesses and nonprofit public interest groups to support full-time employees' participation in standard setting.[125]

Our core concern is not the participation and procedures followed by SSOs per se. We do not think the problem can be solved simply by making SSOs more open and democratic in their processes. The democratic concern engages with the *premise* that the standards we need for AI are properly generated by private actors who are not accountable to the public. SSOs have their origins in nineteenth- and early twentieth-century efforts to coordinate engineers and scientists on common systems of measurement and design choices for physical objects: standards for resistance coils in telegraph systems, screw threads, and chemical and performance characteristics of steel rails for railroads, for example.[126] In these early examples, the benefits of standardization are primarily in terms of market efficiency—promoting comparability of products, interoperability, and markets for parts. Other than the public interest in well-functioning markets, there was little public concern at stake in the standards set for screw threads and steel rails, and hence little need for democratic oversight. Moreover, in those cases in which the public interest was sometimes adjudicated through technical choices, the public interest was uncontested. As Yates and Murphy recount, the dangers associated with steam boilers, which sometimes exploded on riverboats and caused highly publicized deaths, were ultimately addressed with safety standards governing boiler design, materials, construction, and maintenance developed by the private Franklin Institute in 1836.[127] Again, democratic oversight to develop technical standards to avoid boiler explosions in public places seems unnecessary. We can all agree that boilers on riverboats should not explode, just as today we can all agree that airplanes should not fall from the sky.

But the matters being left to technical standard setting in the context of AI governance are anything but purely technical. We are probably well advised to leave to engineers the development of standards that will reduce to minimal the risk that an autonomous vehicle will randomly lose control. But the AI standards that governments are electing globally to leave to private technical standard-setting bodies go far beyond noncontroversial safety standards. They include core human values and choices about the shape of our social and economic lives.

---

124. NAT'L INST. OF STANDARDS & TECH., *supra* note 95, at 8.

125. *See* TIM BÜTHE & WALTER MATTLI, THE NEW GLOBAL RULERS: THE PRIVATIZATION OF REGULATION IN THE WORLD ECONOMY 154–57, 157 n.92 (2011); THE POLITICS OF GLOBAL REGULATION (Walter Mattli & Ngaire Woods eds., 2009); *see also* Walter Mattli & Tim Büthe, *Global Private Governance: Lessons from a National Model of Setting Standards in Accounting*, LAW & CONTEMP. PROBS., Summer-Autumn 2005, at 225, 242 ("Large corporations have the necessary resources, including technical expertise and organizational structure to be actively involved, and industry groups contribute an average of sixty to sixty-five percent of responses to FASB discussion memoranda and exposure drafts."); Veale & Borgesius, *supra* note 13.

126. JOANNE YATES & CRAIG N. MURPHY, ENGINEERING RULES: GLOBAL STANDARD SETTING SINCE 1880, at 19, 22 (2019).

127. *Id.* at 28–29.

They are, fundamentally, not matters on which it is reasonable or appropriate to expect consensus to emerge.

Consider, for example, work from standard-setting bodies with respect to algorithmic bias: the risk that AI systems such as facial recognition or automated decision-making will discriminate against people on protected characteristics such as race or gender. ISO/IEC TR 24027, for example, is a standards document entitled *Bias in AI Systems and AI Aided Decision Making* published by ISO in November 2021.[128] It consists of a technical report that discusses sources of bias in an AI system ("human cognitive bias," "data bias," and "bias introduced by engineering decisions") and lays out a set of statistical metrics to assess bias and methods for treating unwanted bias.[129] The concepts and measures of bias in this document, however, are controversial.[130] They reflect technical approaches to a societal harm that is addressed elsewhere in the economy not through statistical analysis, but through nuanced legal analysis (which can include consideration of statistical evidence) of rich concepts such as disparate impact or intentional discrimination and through adjudication of controversies guided by a history of caselaw or explicit legislation, conducted by specialized regulators, judges and juries, not engineers.[131] These regulators, judges, and juries possess democratic legitimacy because they are ultimately accountable to democratic oversight. Technical standard setting is simply inappropriate as a method of regulating the phenomenon of discrimination, even when it occurs through the operation of a system with important technical elements.

Such ISO standards are only one contribution to the pool of competing alternative standards that industry might adopt and so we could expect pressure for standards over time to evolve more detailed guidance for industry. But market competition between SSOs can only develop more detailed technical guidance. It cannot remedy the democratic deficit. Our model for regulatory markets is precisely intended to achieve this by ensuring that competition over alternative technical approaches to regulating specific attributes and outcomes of AI systems is ultimately accountable to the public via government oversight.

A democratic deficit also limits the capacity for the technical deficit to be met through industry R&D. Businesses have been internally developing technical standards for safe AI. Consider the work being done to ensure powerful generative AI models do not promote harmful outcomes such as fostering misinformation or persuading people to engage in self-harm. The safety of these models is being addressed by developers through a combination of fine-tuning such as reinforcement learning from human feedback (RLHF)[132] and by adversarial testing in which "red teams" attempt to get the model to produce harmful outputs, evading hand-coded and RLHF-generated guardrails.[133] These processes are ideal examples of what we mean by *regulatory technology*. RLHF finetuning involves machine-learning methods, both to capture human values and to achieve the necessary scale to handle the open-ended behaviors of a massive, and massively used, model. Red teaming ideally involves systematic attempts to break a model, including sophisticated attempts using machine-learning tools such as searching for the sometimes, unintuitive prompts that will "jailbreak" a model.[134] These technologies are expensive to research and develop; they require investment and technical sophistication. But that investment and expertise should be pointed ultimately at achieving outcomes that are overseen by governments, not corporate managers, CEOs, and board members. Our regulatory markets proposal achieves that transformation of current regulatory technology development efforts: In our model, RLHF and red-teaming efforts would be, at least in part, shifted to an independent technology sector that answers both to governments (through licensing) and to developers (through competition in the market for the regulatory services developers are required to purchase).

---

128. *ISO IEC TR 24027:2021—Information Technology—Artificial Intelligence (AI)—Bias in AI Systems and AI Aided Decision Making*, INT'L ORG. FOR STANDARDIZATION (Nov. 2021), https://www.iso.org/standard/77607.html [https://perma.cc/ETK4-E6TA] (standard available for purchase).

129. INT'L ORG. FOR STANDARDIZATION, ISO/IEC TR 24027:2021, ARTIFICIAL INTELLIGENCE (AI)—BIAS IN AI SYSTEMS AND AI AIDED DECISION MAKING §§ 6.2–6.3 (2021).

130. *See* Victor Galaz et al., *Artificial Intelligence, Systemic Risks, and Sustainability*, TECH. SOC'Y, Nov. 2021, art. no. 101741, at 1.

131. *Cf.* Deirdre K. Mulligan et al., *This Thing Called Fairness: Disciplinary Confusion Realizing a Value in Technology*, PROCEEDINGS ACM ON HUMAN-COMPUTER INTERACTION, Nov. 2019, art. no. 119, at 1.

Direct government participation has already increased in the setting of technical standards with respect to frontier models. The October 2023 U.S. executive order, E.O. 14110, explicitly directed NIST to develop "guidance and benchmarks for evaluating and auditing AI capabilities" relevant to cybersecurity and biosecurity and for red teaming by developers "to enable deployment of safe, secure, and trustworthy systems."[135] It also directed the Secretary of Energy to develop model evaluation tools and testbeds and model guardrails for capabilities relevant to "nuclear, nonproliferation, biological, chemical, critical infrastructure and energy-security threats."[136] Although this order was rescinded in January 2025, the AI Action Plan issued by the Trump administration in August 2025 recommended that NIST launch standard-setting efforts in specific domains[137] and support the development of an "AI evaluations ecosystem."[138] Similarly, in creating the AI Safety (now Security) Institute in November 2023, the United Kingdom explicitly grounded its mission in the "conviction that governments have a key role to play in providing publicly accountable evaluations of AI systems."[139] Both countries continue to stop short of mandating compli-

132. *See, e.g.*, Paul F. Christiano et al., *Deep Reinforcement Learning from Human Preferences*, ARXIV (Feb. 17, 2023), https://arxiv.org/pdf/1706.03741 [https://web.archive.org/web/20250608010347/http://arxiv.org/pdf/1706.03741].

133. *See, e.g.*, APOSTOL VASSILEV ET AL., NIST AI 100-2E2023, ADVERSARIAL MACHINE LEARNING: A TAXONOMY AND TERMINOLOGY OF ATTACKS AND MITIGATIONS 42 (2024).

134. *See, e.g.*, *id.* at 37.

135. Exec. Order No. 14110, 88 Fed. Reg. 75191, 75196 (Oct. 30, 2023).

136. *Id.*

137. WHITE HOUSE AI ACTION PLAN, *supra* note 97, at 5

138. *Id.* at 10.

139. U.K. DEP'T FOR SCI. INNOVATION & TECH., *supra* note 105, at 7.

ance with these standards or use of the evaluation tools developed by these government actors. Similarly, the E.U. approach, although it largely delegates technical details to standards developed by industry bodies, anticipates that its AI Office will participate in helping to establish technical requirements for generative AI.[140] Much of how this will work out is still unknown. As we will outline below, we believe all these approaches will benefit from a regulatory markets approach.

## D. China

We conclude this review of the AI governance landscape with a context that challenges our discussion of democratic deficits in the approaches we see evolving in the West. China is both a major global player in the production of AI models[141] and an early mover in AI governance. Moreover, although AI policy debates in the West are increasingly framed in terms of geopolitical competition with China—we can't regulate AI because China won't and we will lose the innovation race[142]—a careful look at China's approach to AI regulation both belies this caricature and, importantly, may identify more common regulatory ground than has been appreciated. Moreover, as we will discuss, our proposal for regulatory markets to resolve technical and democratic deficits in Western AI governance needs to account for if and how China might participate in global regulatory markets.

China initially adopted concrete legal requirements for AI systems largely out of concern for how AI-powered recommendations on platforms could disrupt party-state control over online discourse.[143] A 2021 regulation required recommender algorithms to align with existing laws controlling internet news and to "uphold mainstream value[s]."[144] But the law was not limited to these concerns: It also created rights for users to turn off personalized recommendations and specific tags and prohibited systems that caused addiction or unsafe behaviors, unsafe or unfair working conditions (for workers such as food delivery drivers whose jobs relied on AI-based scheduling), and price discrimination.[145] The law created an algorithm registry, requiring disclosure of an algorithm's service provider's name, the type of algorithm, details about the operation of the algorithm, and a self-assessment report focused on safety and security.[146] Subsequent concerns about the impact of deepfakes led to the adoption in 2023 of a regulation on synthetic media, which "required AI providers to watermark AI-generated content and ensure that content does not violate people's 'likeness rights' or harm the 'nation's image.'"[147] Providers of AI-generated content services are required to verify the real identity of users of those services.[148] This "deep synthesis" regulation also requires registration of algorithms that have "public opinion properties or capacity for social mobilization."[149]

---

140. E.U. AI Act, *supra* note 45, arts. 56, 62(3).

141. INST. FOR HUMAN-CENTERED A.I., STANFORD UNIV., ARTIFICIAL INTELLIGENCE INDEX REPORT 3 (2025), https://hai.stanford.edu/assets/files/hai_ai_index_report_2025.pdf [https://perma.cc/JW5B-R4FR].

142. According to Eric Schmidt, former CEO of Google, "China's not busy . . . stopping things because of regulation." Scale AI, *Eric Schmidt (Former Google CEO): A Global Perspective on AI With*, YOUTUBE, at 20:21 (Oct. 6, 2021), https://www.youtube.com/watch?v=CmBpsw1ORQ0; *see Eric Schmidt Cozies Up to China's AI Industry While Warning U.S. of Its Dangers*, TECH TRANSPARENCY PROJECT (Apr. 11, 2024), https://www.techtransparencyproject.org/articles/eric-schmidt-cozies-up-to-chinas-ai-industry-while [https://perma.cc/C59C-PNKL]; *see also* Andrew Tillett, *China Will Win AI Race if Research Paused: Ex-Google Chief,* AUSTL. FIN. REV. (Apr. 6, 2023), https://www.afr.com/politics/federal/china-will-win-ai-race-if-research-paused-ex-google-chief-20230405-p5cy7v.

143. MATT SHEEHAN, CHINA'S AI REGULATIONS AND HOW THEY GET MADE (2023) [hereinafter SHEEHAN, CHINA'S AI REGULATIONS], https://carnegie-production-assets.s3.amazonaws.com/static/files/202307-Sheehan_Chinese%20AI%20gov-1.pdf [https://perma.cc/UJ4W-USQM]; MATT SHEEHAN, TRACING THE ROOTS OF CHINA'S AI REGULATIONS (2024) [hereinafter SHEEHAN, TRACING THE ROOTS], https://carnegie-production-assets.s3.amazonaws.com/static/files/Sheehan_Reverse_Engineering_AI_Gov-UPDATED-1.pdf [https://perma.cc/2Q5C-RJET].

144. SHEEHAN, CHINA'S AI REGULATIONS, *supra* note 143, at 12.

Like Western countries, in the wake of ChatGPT, China began expanding its regulatory approach to encompass the risks of generative AI. In August 2023, the Interim Measures for the Administration of Generative Artificial Intelligence Services came into effect.[150] Provisions of this regulation require providers of generative AI services to "[a]dhere to the core socialist values" and ensure their services do not endanger national security or stability or promote terrorism, extremism, ethnic hatred, discrimination, violence, obscenity, or misinformation.[151] When training models, AI developers must "[t]ake effective measures to prevent discrimination based on ethnicity, religion, country, region, gender, age, occupation, health, etc.,"[152] respect intellectual property rights and anti-monopoly laws, "not infringe upon others' portrait rights [likenesses], reputation rights, honor rights, privacy rights and personal information rights"[153] and "[t]ake effective measures based on the characteristics of the service type to enhance the transparency of generative AI services and improve the accuracy and reliability of generated content."[154] Developers are required to use data and models obtained only from legal sources and to effectively supervise labeling activities.[155] Furthermore, the regulation states that "[t]hose who provide generative artificial intelligence services with public opinion attributes or social mobilization capabilities shall conduct security assessments" and register with the algorithm registry.[156]

---

145. *Id.*

146. *Id.* at 13.

147. SHEEHAN, TRACING THE ROOTS, *supra* note 143, at 1.

148. SHEEHAN, CHINA'S AI REGULATIONS, *supra* note 143, at 13.

149. SHEEHAN, TRACING THE ROOTS, *supra* note 143, at 25.

150. Cyberspace Administration of China, *Interim Measures for the Administration of Generative Artificial Intelligence Services* [国家互联网信息办公室, 生成式人工智 能服务管理暂行办法], (July 13, 2023), https://www.cac.gov.cn/2023-07/13/c_1690898327029107.htm [https://perma.cc/NP8U-KDY8] (click select "English" from Google Translate to view an English language version).

151. *Id.* art. 4(1).

152. *Id.* art. 4(2).

153. *Id.* art. 4(3)(iv).

154. *Id.* art. 4(3)(V).

155. *Id.* arts. 7(1), 8.

156. *Id.* art. 17.

China's AI regulations share features of both the E.U. and U.S./U.K. approaches. On the one hand, like the E.U. AI Act, China's regulations set requirements about how AI algorithms (more generally, models or systems) are built and how they behave, regardless of sector. Moreover, although few would characterize the Chinese approach to AI governance as democratic in the sense in which we used that term above, the Chinese government is clearly making political choices about how AI is developed and deployed. They are relying on the mechanism of mandatory law rather than voluntary industry-drafted standards and guidelines to regulate. On the other hand, as in the United States and the United Kingdom, Chinese regulations are emerging from sectoral regulators (most importantly, the Cyberspace Administration of China, which is responsible for regulating the Chinese internet) and at least initially have focused on specific challenges like recommendation systems and synthetic media.

Also, like their Western counterparts, China's regulators have faced the challenge of establishing requirements that operationalize legal controls with sufficient technical sophistication, specificity, and capacity for rapid adaptation to technological developments. China, too, has turned to technical standards bodies to generate detailed technical requirements. The 2023 Interim Measures for generative AI, for example, have been translated into specific tests and benchmarks by the Standardization Administration of China, an organization authorized by the government to coordinate technical standards in China and represent China in international standards bodies like the ISO.[157] This process has moved quickly to create fairly concrete operational requirements. For example, the standard requires that developers establish a comprehensive keyword library of at least 10,000 words, a comprehensive generated content question bank of no fewer than 2,000 questions, and test question banks of at least 500 questions each of first, questions the model should refuse to answer and second, questions the model should not refuse to answer.[158] The standard then sets numerical limits on the frequency with which a sampling of the model conducted using these comprehensive libraries and question banks generates acceptable responses.[159] For example, the refusal rate on questions the model should refuse must not be lower than ninety-five percent and the refusal rate on questions the model should not refuse must not be greater than five percent.[160] While these operational standards appear to provide clear guidance for compliance, the effect they will have on the quality and cost of the systems remains unknown.

157. *E.g.*, NAT'L TECH. COMM. 260 ON CYBERSECURITY OF STANDARDIZATION ADMIN. OF CHINA, TECHNICAL DOCUMENTATION OF NATIONAL TECHNICAL COMMITTEE 260 ON CYBERSECURITY OF STANDARDIZATION ADMINISTRATION OF CHINA: BASIC SAFETY REQUIREMENTS FOR GENERATIVE ARTIFICIAL INTELLIGENCE SERVICES (2024), https://cset.georgetown.edu/publication/china-safety-requirements-for-generative-ai-final/ [https://web.archive.org/web/20250604204600/https://cset.georgetown.edu/publication/china-safety-requirements-for-generative-ai-final/] (Click "Download Full Translation").

158. *Id.* at 10–11.

159. *See id.* at 14.

160. *Id.* at 14–15.

A key distinction between Chinese and Western regulation of AI remains: Chinese law requires private companies to maintain a Communist Party apparatus internally that gives the party-state direct visibility into, and perhaps increasingly a say in, the operations of and technology being developed by AI companies in China.[161] Whereas Western governments must develop regulatory methods that respect the boundaries and autonomy of the private companies building and deploying AI and which can achieve democratic legitimacy and oversight under that constraint, Chinese regulators operating in a one-party regime are able to intervene directly in the development and deployment of the technology and, if the faster rate at which China has produced technically detailed AI regulation in some domains is enough to go on, more rapidly.[162]

## III. REGULATORY MARKETS

Our proposal for regulatory markets builds on existing work in regulatory theory. In this Part, we first review this existing work and then lay out how regulatory markets would function as an evolution of new governance and regulatory intermediary theory and how it would apply to AI.

### A. New Governance and Regulatory Intermediary Theory

The challenge of regulatory innovation to address technological change is not a new one. In recent decades, regulatory theorists have proposed "new governance" techniques to supplement or displace traditional command-and-control forms of regulation.[163] New governance responds to the call for more "agile" governance[164] in the face of rapid change and high levels of complexity.

These new governance techniques fall into three large groups. Performance-based regulation (also called outcomes-based or principles-based regulation) specifies results (sometimes expressed as metrics, sometimes as principles) that regulated entities must achieve but does not specify how to achieve those results.[165] Management-based regulation (also called process-oriented, risk-based, or enforced self-regulation) requires firms to evaluate the risks generated by their business and to develop their plan for how those risks will be managed. Plans might need approval from the government or a third-party certification agency.[166] Meta-regulation embeds these new governance techniques in a system in which both regulated entities and government regulators continually learn from experience to update required processes and outcomes.[167]

---

161. YUKYUNG YEO, VARIETIES OF STATE REGULATION: HOW CHINA REGULATES ITS SOCIALIST MARKET ECONOMY (2020); Scott Livingston, *The New Challenge of Communist Corporate Governance*, CTR. FOR STRATEGIC & INT'L STUD. (Jan. 2021), https://csis-website-prod.s3.amazonaws.com/s3fs-public/publication/210114_Livingston_New_Challenge.pdf [https://perma.cc/Z66L-L62U].

162. Some observers, however, see China's regulatory approach as an effort to appease industry, offering few actual protections to the Chinese public. Angela Huyue Zhang, *The Promise and Perils of China's Regulation of Artificial Intelligence*, 63 COLUM. J. TRANSNAT'L L. 1, 1 (2025).

163. *See, e.g.*, BRAITHWAITE & DRAHOS, *supra* note 10; Carrigan & Coglianese, *supra* note 10.

164. *See, e.g.*, Glob. Agenda Council on the Future of Software & Soc'y, *A Call for Agile Governance Principles*, WORLD ECON. F. (2016), https://www3.weforum.org/docs/IP/2016/ICT/Agile_Governance_Summary.pdf [https://perma.cc/H278-GGYK]; WORLD ECON. F., AGILE GOVERNANCE: REIMAGINING POLICY-MAKING IN THE FOURTH INDUSTRIAL REVOLUTION (2018), https://www3.weforum.org/docs/WEF_Agile_Governance_Reimagining_Policy-making_4IR_report.pdf [https://perma.cc/THQ8-DAUW]; WORLD ECON. F., GLOBAL TECHNOLOGY GOVERNANCE: A MULTISTAKEHOLDER APPROACH (2019), https://www3.weforum.org/docs/WEF_Global_Technology_Governance.pdf [https://perma.cc/6JXP-6QCB]; WORLD ECON. F., AGILE REGULATION FOR THE FOURTH INDUSTRIAL REVOLUTION: A TOOLKIT FOR REGULATORS (2020), https://www3.weforum.org/docs/WEF_Agile_Regulation_for_the_Fourth_Industrial_Revolution_2020.pdf [https://perma.cc/UTM4-PHKN].

The move to new modes of regulation has been fostered by the perception that traditional approaches inhibit both efficiency and innovation in the achievement of regulatory goals. The theory of new governance approaches is that government should harness the expertise and cost-minimizing incentives of industry itself in the pursuit of politically established goals such as a safe food supply, reduced pollution, or stable financial systems.[168]

A critical feature of new governance methods is increased reliance on what Kenneth W. Abbott, David Levi-Faur, and Duncan Snidal call *regulatory intermediaries*.[169] They define an intermediary as "any actor that acts directly or indirectly in conjunction with a regulator to affect the behavior of a target."[170] Intermediaries can include private actors (such as for-profit companies supplying auditing and certification services), civil society organizations (such as NGOs supplying fair trade standards) or government agencies (such as independent national bodies established to oversee human rights compliance).[171] In their regulatory intermediary theory (RIT), instead of a direct relationship between a regulator and a target,

**R ⟹ T,**

they posit a relationship that interposes a third-party intermediary, making the relationship between regulator and target indirect:

**R ⟹ I ⟹ T.**[172]

In this model, regulators integrate regulatory intermediaries into a governance scheme as a means of recruiting the intermediaries' greater capacity to implement, monitor, or enforce rules and the role they can play in providing feedback about rule performance for purposes of rule revision.[173] Intermediaries may also play a role in "soft" RIT regulation by generating consensus-based nonbinding standards (such as fair trade or forest stewardship standards) that help direct an industry in ways desired by a government.[174] This is the sense of regulation defined by Julia Black: "sustained and focused attempts to change the behavior of others in order to address a collective problem or attain an identified end or ends, usually through a combination of rules and norms and some means for their implementation and enforcement, which can be legal or non-legal."[175] The more actors involved as intermediaries, the more *polycentric* a regulatory regime is; the greater the reliance on non-state actors, the more *decentered* it is.[176] The approach seen in existing responses to the AI governance challenge clearly reflect a decentered regime with substantial reliance on regulatory intermediaries—namely nonstate standard setting bodies and private auditing and certification
services.

---

165. *See* Coglianese et al., *supra* note 36; Gilad, *supra* note 36; Braithwaite, *The Essence of Responsive Regulation*, *supra* note 36; Peter J. May, *Performance-Based Regulation*, *in* HANDBOOK ON THE POLITICS OF REGULATION 373 (2011); AYRES & BRAITHWAITE, *supra* note 36.

166. *See* Gilad, *supra* note 36; Braithwaite, *The Essence of Responsive Regulation*, *supra* note 36, at 508; Coglianese et al., *supra* note 36, at 718; AYRES & BRAITHWAITE, *supra* note 36, at 106.

167. Sharon Gilad, *It Runs in the Family: Meta-Regulation and Its Siblings*, 4 REGUL. & GOVERNANCE 485, 485 (2010).

168. *See* CHRISTINE PARKER, THE OPEN CORPORATION: EFFECTIVE SELF-REGULATION AND DEMOCRACY 14 (2002).

169. Kenneth W. Abbott et al., *Theorizing Regulatory Intermediaries: The RIT Model*, 670 ANNALS AM. ACAD. POL.& SOC. SCI. 14 (2017).

170. *Id.* at 19 (emphasis omitted).

171. *Id.* at 15.

172. *Id.* at 26.

Our model extends the regulatory intermediary model in a new critical direction. Specifically, we focus on the goal of building a *regulatory market* not merely to recruit the expertise or cost-efficiency of non-state intermediaries but, more importantly, to attract investment of human and financial capital into the design and development of true regulatory *technologies* that can keep pace with AI. This is our proposed response to the technical deficit we see in governments' efforts to regulate AI directly. We also propose, however, to structure regulatory markets in a framework that addresses the democratic deficit of existing approaches that lie too far on the industry/self-governance end of the regulatory spectrum.

## B. The Regulatory Markets Model

There are three principal actors in our regulatory markets model: the targets of regulation, private regulators, and governments. *Targets* are businesses and other organizations that governments seek to regulate. In the AI context, these are the companies or organizations building and deploying AI or integrating AI into their products, services, or systems. *Private regulators* are for-profit and nonprofit organizations that develop and supply regulatory services, which they compete to sell to targets. *Governments* require targets to purchase regulatory services (entering into a regulatory contract with a private regulator) and directly regulate the market for regulatory services, ensuring it operates in the public interest. Private regulators would gain their authority to regulate via the regulatory contract with the target and authorization from governments to collect fines or impose requirements on the targets that submit to their regulatory system.

---

173. *See id.* at 29–30.

174. *See id.* at 26.

175. Julia Black, *Constructing and Contesting Legitimacy and Accountability in Polycentric Regulatory Regimes*, 2 REGUL. & GOVERNANCE 137, 139 (2008).

176. *See id.* 139–40; Julia Black, *Decentring Regulation: Understanding the Role of Regulation and Self-Regulation in a 'Post-Regulatory' World*, 54 CURRENT LEGAL PROBS. 103, 103–04 (2001).

(See Figure 2 for an illustration of how this model compares schematically to traditional command-and-control regulation.) We discuss each of the elements of this model in turn.

### 1. *Private Regulatory Services: Regulatory Technology*

Private regulators could employ conventional means of regulation: writing text-based rules, monitoring for compliance, and penalizing violations. But the private regulator might also develop *regulatory technologies* that directly control or shape the business decisions of the targets it regulates. And indeed, this is a primary goal: encouraging investment in novel methods of aligning the behavior of targets with desired outcomes, overcoming the technical deficit of direct government regulation.

The concept of regulatory technology is not new. "RegTech" first appeared as a term in financial regulation and in that context has been defined as "the use of technology, particularly information technology (IT), in the context of regulatory monitoring, reporting, and compliance."[177] The development of regulatory technology in this context took off as financial transactions became increasingly digitized and in the wake of the 2008 global financial crisis, as countries around the world significantly increased reporting and compliance requirements for financial institutions.[178] This regulatory shift created an increased demand from financial institutions for automated processes to ensure compliance with a raft of new rules for stress testing, risk assessments, reporting of derivative transactions, and more.[179] This demand was largely met by private sector innovations in compliance technologies, making use of high volume data scraping, big data analysis, and machine learning.[180] Additionally, the vision of a new regime in which regulators themselves would deploy technology to implement financial oversight emerged—with the Chief Economist of the Bank of England articulating a "dream" in 2014 of a Star Trek future in which regulators have a bank of monitors "tracking the global flow of funds in close to real time."[181] Financial scholars at the time envisioned a role for private sector vendors and academic researchers to supply governments with new digital tools and techniques for achieving that vision[182] and by 2021 some experiments in this

177. Douglas W. Arner et al., *FinTech, RegTech, and the Reconceptualization of Financial Regulation* 37 NW. J. INT'L L. & BUS. 371, 373 (2017); Veerle Colaert, *'Computer Says No'—Benefits and Challenges of RegTech*, *in* ROUTLEDGE HANDBOOK OF FINANCIAL TECHNOLOGY AND LAW 431 (Iris Chiu & Gudula Deipenbrock eds., 2021) [hereinafter ROUTLEDGE HANDBOOK].

178. *See generally* Arner et al., *supra* note 177; Colaert, *supra* note 177.

179. *E.g.*, Arner et al., *supra* note 177, at 389.

180. *See generally* Kenneth A. Bamberger, *Technologies of Compliance: Risk and Regulation in a Digital Age*, 88 TEX. L. REV. 669 (2010); Luca Enriques, *Financial Supervisors and RegTech: Four Roles and Four Challenges*, REVUE TRIMESTRIELLE DE DROIT FINANCIER [CORP. FIN. & CAPITAL MARKETS L. REV.], Dec. 2017, at 53 (Fr.).

181. Arner et al., *supra* note 177, at 373 (quoting Andy Haldane, Chief Economist, Bank of Eng., Speech at the Maxwell Fry Annual Global Finance Lecture: Managing Global Finance as a System, Birmingham University 10 (Oct. 29, 2014), https://www.bankofengland.co.uk/-/media/boe/files/speech/2014/managing-global-finance-as-a-system.pdf).

182. Arner et al., *supra* note 177.

domain had begun.[183] Further development of the concept also envisioned the translation of text-based legal requirements into machine-readable and executable code.[184]

Here are a few examples of regulatory technologies that private regulators might develop in the AI domain:

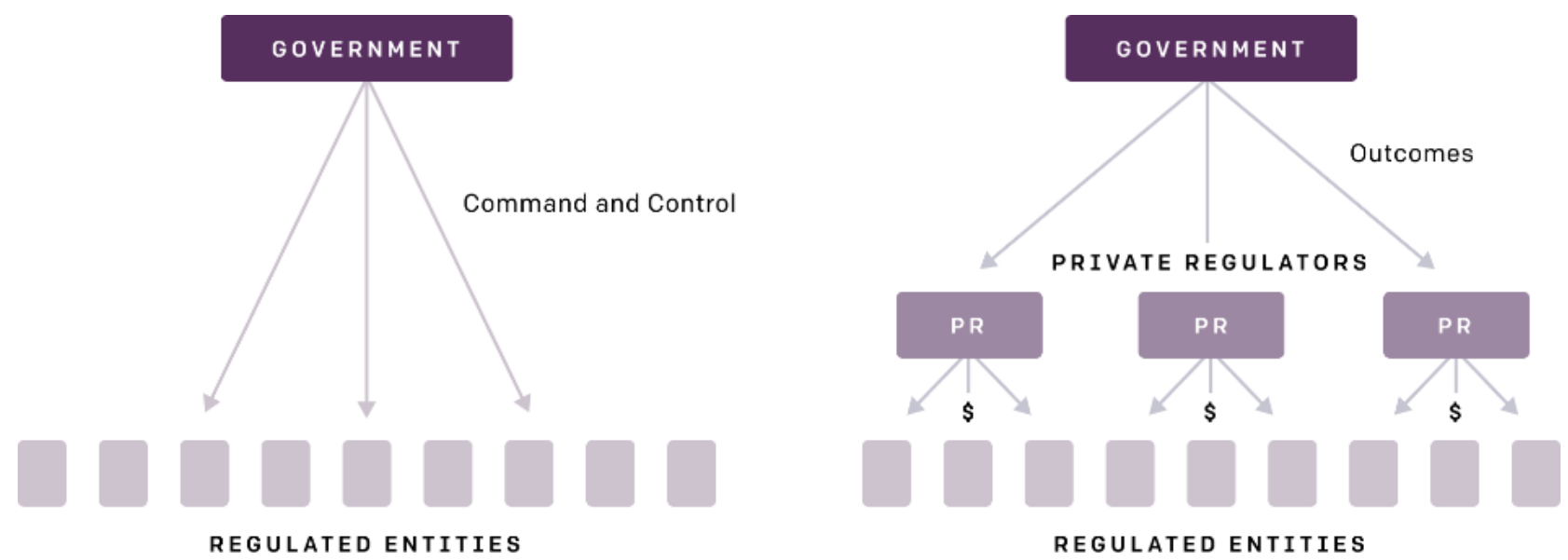

**Figure 2. Examples of Regulatory Technologies.** In conventional regulation, shown on the left, the government directly regulates entities. With regulatory markets, shown on the right, a private regulator directly regulates the targets that have purchased its regulatory services, subject to oversight by government to ensure regulators are achieving outcomes set by government. It does this by developing regulatory procedures, requirements, and technology.

- A private regulator of self-driving cars might require self-driving car companies to allow the regulator access to data produced by the vehicles and then use machine learning (ML) to detect behaviors that raise the risk of accidents beyond thresholds set by the regulator. The private regulator might bring these to the attention of the target and require risk assessment, or it might develop technology that allows the regulator to modify the algorithms or data sources used by the target's vehicles.

- A private regulator in the banking industry might require a bank using ML to analyze customer data and develop new products to implement differential privacy techniques[185] to minimize the likelihood that a customer is harmed by the use of their data. The regulator could prescribe the specific techniques/algorithms to use; or it could establish a procedure for the banks that it regulates to propose techniques that survive tests conducted by the regulator.

- A private regulator of drones equipped with facial recognition systems might require developers and manufacturers to implement particular cy-

183. Paola Chirulli, *FinTech, RegTech and SupTech: Institutional Challenges to the Supervisory Architecture of the Financial Markets*, ROUTLEDGE HANDBOOK, *supra* note 177, at 447.

184. Colaert, *supra* note 177, at 434.

185. *See, e.g.*, Cynthia Dwork & Aaron Roth, *The Algorithmic Foundations of Differential Privacy*, *in* 9 FOUNDS. & TRENDS THEORETICAL COMPUT. SCI. 211 (2014).

> bersecurity features to ensure their models are not discoverable by malicious users and conduct algorithmic audits of system accuracy across demographic groups. The regulator might also create systems that enable people to raise flags about drone behavior to detect malicious or discriminatory use.

Regulatory techniques developed by private regulators might include hardware, risk assessment tools, information processing systems, conflict or complaint management procedures, red-teaming protocols and benchmarks, and so on.

2. *Licensing: Outcome-Based Regulation*

To participate in the market by selling regulatory services to targets, private regulators must be first licensed by the government in the jurisdictions in which they wish to operate. In any given domain, multiple regulators are licensed so that they compete to provide regulatory services to targets. Targets must choose a regulator, but they have the capacity to choose, and switch, regulators. They do so by comparing across regulators in terms of the cost and efficiency of the services provided by regulators.

Private regulators do not compete, however, on the quality of their regulatory services, that is, the extent to which they achieve public goals. This is because to obtain and maintain a license, regulators must demonstrate their regulatory approach achieves *outcomes* that are mandated by government. Outcomes are metrics or principles set through the bureaucratic processes of the public sector. They are the mechanism by which the delegation of regulatory oversight of target to private actors is made legitimate, overcoming the democratic deficit.[186]

For example,

- In the self-driving car context, governments could set metric thresholds for accident rates or traffic congestion. They could establish principles for private regulators such as maintaining public confidence in road safety.
- In the banking industry, governments could set metric thresholds for access to credit by consumers that must be met by licensed private regulators. They could establish principles such as traceability of transactions and maintenance of confidence in the stability of financial markets.
- In the context of facial recognition use in drones, governments might establish metric thresholds for the likelihood that software could be accessed by malicious users. They could establish principles such as realistic consumer consent to recognition and/or metrics to ensure comparable likelihoods of identification across different demographic groups is local population.

The central innovation of regulatory markets beyond existing new governance models is that it calls for a shift by government to establishing the goals of regulation rather than the methods of achieving those goals (the concept behind performance-based regulation), but using methods developed by independent

186. *See* Abbey Stemler, *Regulation 2.0: The Marriage of New Governance and Lex Informatica*, 19 VAND. J. ENT. & TECH. L. 87 (2016).

private regulators who are themselves regulated by governments (rather than leaving the methods to achieve goals to regulated entities, as we see in performance-based regulation). The regulation of private regulators would occur through a combination of upfront evaluation of the capacity for a regulator's system to satisfy government goals and ongoing auditing and oversight: measurement of outcome metrics and assessment of the achievement of principles by the private regulator.

For example, in the self-driving car setting, governments may develop techniques to track accident and congestion rates and assess the contribution of a particular regulator to excessive accidents or congestion. In banking, governments could conduct periodic audits of random samples of transactions from the targets of a particular regulator to determine the incidence of money laundering. In drones, governments might stress test a regulator's procedures by employing adversarial efforts to infiltrate algorithms or data.

Regulators that fail to pass the tests set by governments would risk having their licenses suspended, conditioned, or revoked. This requires governments to develop the regulatory models and technical expertise needed to effectively evaluate the outcomes achieved by private regulators and to ensure that the threat of losing a license for poor performance is realistic. Making sure this threat is realistic also requires capacity to ensure that the market for private regulators is competitive, that there is sufficient scale in a given domain to support multiple regulators (possibly restricting the share of the target market that a given regulator can service), and that targets have the capacity to switch regulators with relative ease. This obtains the benefits of competition between regulators, spurring them to invest in developing more effective and less costly means of achieving regulatory objectives.

Our model updates the regulatory technology concept as it has been developed to date in the context of financial regulation. In the financial regulation context, regulatory technology has been deployed to digitize existing regulatory components: translating legal requirements into computer code and automating compliance procedures.[187] The role of third-party private sector vendors (or internal government or business engineering staffs) has been limited to fulfilling the demand for digitization and automation. In our model for AI, we envision expanding the role for private sector vendors to include the translation not (only) of text-based legal rules into computer code but, more fundamentally, the translation of government-supplied outcomes into vendor-supplied technical requirements. We also envision a transformation of the role of government, from the creation of specific technical requirements to regulate target companies to the robust oversight of this translation and implementation of regulatory requirements by private sector providers.

### C. Regulatory Markets for Red-Teaming Frontier Models

We discussed above the ways in which regulatory technologies are already being developed to address safety concerns for frontier models, emerging from

187. Colaert, *supra* note 177, at 434.

active research efforts within AI companies. But these efforts suffer from the democratic deficit: The criteria for evaluating the value and efficacy of these technologies are being set under corporate governance within private companies. We are advocating for moving at least some of these efforts into a sector of independent technology firms that are answerable both to governments in the form of securing and maintaining licenses and to developers of AI in the form of market competition to secure regulatory services contracts.

In this Section, we sketch in more detail how regulatory markets could work in the context of red-teaming frontier models. Red teaming is a methodology that has been already developed within AI companies such as OpenAI and Anthropic, meaning it is in demand by technology companies already and is a reasonable domain in which to expect regulatory technology to emerge and advance with further investment. Regulatory technology in this domain includes both process elements—such as the selection of testers and tests—and technical elements—such as the training of models that produce sequences of apparently meaningless characters that when appended to a user query circumvent safety guardrails in public-facing models such as ChatGPT.[188] Red teaming has also quickly gained policy currency with governments and industry observers. The U.S. executive order and the E.U. AI Act both reference adversarial testing of frontier general-purpose models. E.O. 14110 explicitly defines "red-teaming,"[189] directs NIST to establish "appropriate guidelines, including appropriate procedures and processes, to enable developers of AI, especially of dual-use foundation models, to conduct AI red-teaming tests,"[190] and requires any developer of a dual-use foundation model to inform the Secretary of Commerce about performance of the model on red-team tests.[191] With this heightened policy attention, it has also become clear that this a domain where it will be critical for testing to be carried out by independent entities, for governments and the public to trust testing reports. AI developers are evolving from using entirely internal teams[192] to retaining the services of nonprofit alignment research groups such

188. *See, e.g.*, Andy Zou et al., *Universal and Transferable Adversarial Attacks on Aligned Language Models*, ARXIV (July 28, 2023), https://arxiv.org/pdf/2307.15043v1 [https://web.archive.org/web/20250614131426/https://arxiv.org/pdf/2307.15043v1].

189. As E.O. 14110 states, "The term 'AI red-teaming' means a structured testing effort to find flaws and vulnerabilities in an AI system." Exec. Order No. 14110, 88 Fed. Reg. 75191, 75194 (Oct. 30, 2023).

190. *Id.* at 75196.

191. *Id.* at 75197.

192. *See, e.g.*, Anthropic, *Red Teaming Language Models to Reduce Harms: Methods, Scaling Behaviors, and Lessons Learned*, arXiv (Nov. 22, 2022), https://arxiv.org/pdf/2209.07858 [https://web.archive.org/web/20250616152005/http://arxiv.org/pdf/2209.07858].

as METR[193] and Apollo Research,[194] and consulting firms like Gryphon Scientific.[195]

We focus in this example on a specific policy objective, namely reducing the risk that frontier models are misused to create biological weapons. In a regulatory markets model, governments would first announce a time frame in which they expected to license independent red teaming and evaluation companies and a date by which developers of frontier models (whether open- or closed-source) would be required to enter into fee-based contracts with a licensed red-teaming company. They would then establish licensing criteria. Understanding that we are still in the very early days of understanding how frontier models might, now or soon, increase the capacity for bad actors to create biological weapons, and how to effectively identify such vulnerabilities, we could anticipate that the initial licensing scheme for red-teaming providers would consist of establishing an expert group. This group would assess red-teaming companies against a general principle such as "warrants high confidence that a model is not vulnerable to state-of-the-art adversarial efforts to increase baseline capabilities among non-state actors to produce category A, B or C bioterror toxins as classified by the U.S. Centers for Disease Control."[196] Judgments about what counts as "high confidence," "not vulnerable," or "state-of-the-art" would be left to this expert group to assess on an ongoing basis. This expert group could provide early guidance to companies that anticipate seeking a license under this general principle and then reach initial determinations based on submissions from the first crop of applicants. What it takes to meet this licensing requirement would then evolve as the expert group engaged in, perhaps initially relatively frequent, evaluations of licensee systems, technology, and performance. Over time, we would expect this general principle to be elaborated with either additional qualitative requirements or specific metrics. The essence of the oversight, however, is that the licensing agency composed of this expert group will focus on outcome

193. METR was founded by Beth Barnes and spun out from the evaluations team at the Alignment Research Center, founded by Paul Christiano. *About METR*, METR, https://metr.org/about [https://perma.cc/E6AL-8KKE]; *Team*, ALIGNMENT RSCH. CTR., https://www.alignment.org/team/ [https://perma.cc/2YCL-AJDJ]. Both Barnes and Christiano previously worked on AI alignment and evaluation at OpenAI. Christiano is now the head of AI Safety for the U.S. AI Safety Institute within NIST. Press Release, U.S. Dep't of Commerce, U.S. Commerce Secretary Gina Raimondo Announces Expansion of U.S. AI Safety Institute Leadership Team (Apr. 16, 2024), https://www.commerce.gov/news/press-releases/2024/04/us-commerce-secretary-gina-raimondo-announces-expansion-us-ai-safety [https://perma.cc/QBB4-TN4X].

194. APOLLO RESEARCH, https://www.apolloresearch.ai/ (last visited Sept. 4, 2025).

195. Gryphon Scientific, a scientific evaluation and consulting firm acquired by Deloitte in April 2024, was retained in 2023 by Anthropic to do analysis and red teaming of how its systems could potentially accelerate efforts to build biological weapons, as outlined in a letter from the Gryphon Scientific's executive chair to Senator Chuck Schumer. Written Statement of Executive Chair of Gryphon Scientific Rocco Casagrande at the U.S. Senate AI Forum: Risk, Alignment and Guarding Against Doomsday Scenarios (Dec. 6, 2023), https://www.schumer.senate.gov/imo/media/doc/Rocco%20Casagrande%20-%20Statement.pdf [https://web.archive.org/web/20250621222039/https://www.schumer.senate.gov/imo/media/doc/Rocco%20Casagrande%20-%20Statement.pdf].

196. *See, e.g.*, Tamar Berger et al., *Toxins as Biological Weapons for Terror—Characteristics, Challenges and Medical Countermeasures: A Mini-Review*, 2 DISASTER & MIL. MED., 2016, art. no. 7, at 1.

measures, namely accomplishing that initial objective of achieving high confidence that frontier models do not significantly increase bioterror threats from non-state actors. Driven by the market opportunity created by licensing regimes, we anticipate that groups that are currently operating as nonprofit research entities would either gain significant new philanthropic funding (along the lines of the multimillion dollar levels of support OpenAI obtained when it was founded in 2015[197]) to enable them to scale up organizationally to actively compete for business[198] or would incorporate as for-profit entities and seek investment capital as startups.

Figure 3 shows how red teaming would work in a regulatory market.

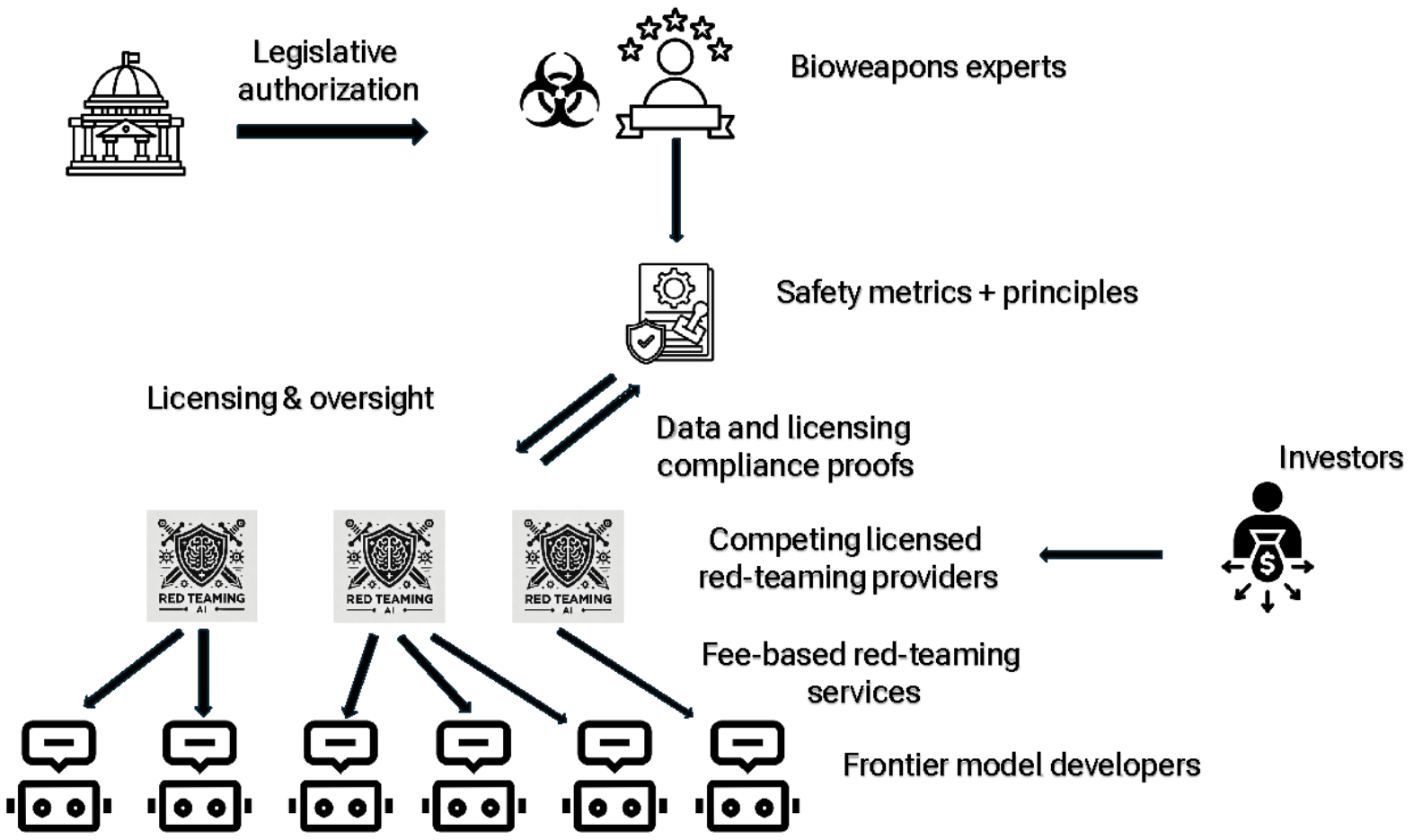

**Figure 3. Red-Teaming Frontier Models for Bioweapon Risk in a Regulatory Market**

## IV. OPPORTUNITIES, LIMITATIONS, AND RISKS

The regulatory markets model is a natural evolution of the widespread use of regulatory intermediaries, particularly in fast-moving and complex contexts, and the development of regulatory technology supplied by private sector vendors in the financial sector. We do not suggest that it is the right solution for all AI settings; we propose it as an addition to the AI governance toolkit. But we believe that it is a critical direction in which AI governance needs to go in order to meet the AI governance challenge of ensuring that regulatory technology evolves at a comparable rate to AI technology itself. This is the key motivation behind the market element in this model: the need to match the market incentive

197. *Our Structure*, OPENAI, https://openai.com/our-structure/.

198. It is important to remember that large nonprofit entities compete actively for business in competitive conditions, such as universities and hospitals.

to invest in AI advances with a market incentive to invest in AI regulation. We do not see any evidence that existing approaches are capable of filling this need. Even if the public sector could compete in the market for technical talent and recruit top engineers into government jobs, which is widely seen as an obstacle to effective AI regulation,[199] publicly employed engineers would not face the incentives and enjoy the supports needed to innovate publicly owned regulatory technologies. This is just not what the public sector is good at. At the same time, the private incentive within AI companies to invest in regulatory technologies is currently unmoored from democratic oversight: Investments are being made to meet corporate goals. These corporate goals are only loosely connected to public goals, given the lack of concrete AI standards and the lack of visibility the public has into how corporate AI systems are functioning.[200]

In this Part, we consider opportunities, limitations, and risks presented by this model.

## A. Opportunities: Meeting the Challenges of Globalization and Regulatory Disruption

The core opportunity presented by the use of regulatory markets is the fostering of increased investment in regulatory technologies that are as sophisticated and agile as the underlying AI technologies they regulate. As many commentators warned after the release of ChatGPT, we are facing a major imbalance in our investment portfolio as a society: We are investing billions in building evermore powerful AI capabilities but very little in building comparably powerful AI governance tools.[201] As we have emphasized, the speed and complexity of AI technologies is almost certain to require other AI technologies to ensure models and systems are producing the outcomes that benefit society. We need, therefore, to harness private-sector incentives to build these regulatory technologies to attract investment capital and technical expertise to the challenge of regulation. It will not be enough to recruit technical experts to government; governments are not designed to take the risks, financial and otherwise, needed to innovate and deliver core technologies.[202] We need to create private-sector opportunities for those currently working within technology companies—building AI systems and developing internal AI safety and oversight techniques—to establish their own businesses with a vision and mission to advance

199. *Cf.* Michael Guihot et al., *Nudging Robots: Innovative Solutions to Regulate Artificial Intelligence*, 20 VAND. J. ENT. & TECH. L. 385, 385 (2017).

200. This problem is partly what gave rise to China's AI registry. *See* SHEEHAN, TRACING THE ROOTS, *supra* note 143. The red-teaming example discussed above could help create investment in regulatory technology and bridge this visibility gap.

201. The Georgetown University Emerging Technology Observatory calculated in 2024 that only two percent of research in AI is focused on AI safety. *AI Safety*, EMERGING TECH. OBSERVATORY, https://almanac.eto.tech/topics/ai-safety/ [https://web.archive.org/web/20250107082722/https://almanac.eto.tech/topics/ai-safety/]. Leading researchers have called for one-third of research funding in industry and academia to be devoted to AI safety. Yoshua Bengio et al., *Managing Extreme AI Risks Amid Rapid Progress*, 384 SCIENCE 842, 844 (2024).

202. HADFIELD, *supra* note 9.

our collective ability to ensure AI systems do not cause individual harm or society-wide disruption and that they do not fall into the hands of malicious users. Moreover, we need to create the opportunities for venture capital to place multiple bets on which technologies and business models in this regulatory sector will succeed.

We also see two other key advantages of this regulatory approach: (1) the capacity for regulatory markets to address the challenge of globalization—stemming from the fact that most AI technologies are deployed in ways that easily transcend jurisdictional boundaries—and (2) the fundamental challenge of AI's potential disruption of almost all regulatory goals.

The expansion of trade with later twentieth-century globalization and the increasing use of digital platforms heightened the call for harmonization of regulations across jurisdictions.[203] AI technologies raise the pressure for harmonized standards, as AI is increasingly trained on massive and global data flows. Precisely because AI technologies are general purpose technologies, which can be expected to disrupt almost all sectors of the economy, however, harmonization of regulation across jurisdictions is often a pipe dream: The world's legal jurisdictions are too varied and dynamic, and protective of their sovereignty, to give up rulemaking to supranational international bodies in everything from health, education, and policing to financial markets and advertising standards. What is deemed fair and safe in one jurisdiction will not easily translate around the globe and political convergence is unlikely. Again, in the spirit of cautionary tales, a global effort begun in 1992 to settle on harmonized regulations for medical devices failed after twenty years to produce a harmonized regime, a core reason being the inability to coordinate the legislative changes need to implement regulatory proposals.[204]

Regulatory markets offer a more promising mechanism for achieving the underlying goal of harmonization, which is not to require governments to agree on regulations in international bodies but rather to reduce the burden on companies operating at global scale of complying with multiple regulatory regimes. As we envision the model, regulatory markets are global markets. Multiple private regulators, ideally, are licensed by multiple governments, each implementing their own outcome requirements. Consider a global market for the regulation of facial recognition technologies to ensure these technologies operate in similar ways for different demographic groups. Suppose that seven private regulators offer this regulatory service. Regulators 1, 2, and 3 use auditing techniques focused on ensuring that the training data used to build a facial recognition system is demographically representative. Regulators 4, 5, and 6 use statistical tests run on audited samples of the facial recognition technology in operation. Regulator 7 employs human review panels—like juries—to adjudicate in qualitative terms whether test cases show fair treatment of different demographic groups. Each country assesses each regulator to determine if the regulator meets the country's outcome goals—which could be evaluated using quantitative or qualitative

203. *Id.*

204. Toshiyoshi Tominaga, *The ICH, the GHTF, and the Future of Harmonization Initiatives*, 47 THERAPEUTIC INNOVATION REGUL. SCI. 572 (2013).

methods. Country A could license all 7 regulators to provide regulatory services in its jurisdiction. Country B could license only regulators 1 to 6, lacking confidence in the qualitative approach of regulator 7. Country C could license only regulators 4 to 7, lacking confidence in ex ante data controls to achieve representativeness. All three countries could have different specific demographic goals—and licensed regulators would be required to implement country-specific demographic requirements. But facial recognition technology providers could choose their regulator, based on which jurisdictions they want to access. Provider X could opt for Regulator 1, and design its technology subject only to the training data requirements of Regulator 1—this would give it access to Country A and B. Provider X might need to provide different specific demographic guarantees for its training data in Country A than in Country B. But it would not have to also aim to satisfy ex post statistical tests or qualitative assessments like those designed by regulators 4 to 7. Meanwhile, Provider Y could opt for Regulator 4, which is licensed in all three countries but which requires Y to satisfy specific statistical tests that could vary from country to country. Provider Z could opt for Regulator 7, foregoing operations in Country B; or it could engage Regulator 6 as well to meet Country B requirements if Country B was lucrative enough and the burden of a different ex post test of its systems was not too great. Figure 4 shows how this might work.

**Figure 4. Own Demographic Outcome Targets.** Global market for licensed private regulators of commercial facial recognition providers. Here, seven private regulators employ three different regulatory technologies (auditing of training data, statistical audits of outputs, and human jury audits of outputs). Countries choose which private regulators to license to provide regulatory services in their jurisdiction, based on their own jurisdiction's demographic outcome targets and their confi-

> dence in the different regulatory technologies used by private regulators. Commercial facial recognition companies choose one (or more) regulators to gain access to those jurisdictions. In this example, Company gains access to Canada and Japan by complying with Regulator 1's audits of training data; Company Y gains access to Canada, Japan, and the UAE by complying with Regulator 4's statistical audits of outputs; and Company Z gains access to Canada and the UAE by complying with Regulator 7's innovation of a system of jury audits of outputs.

In this way, providers achieve the benefits of harmonization—they are subject to only one or a small number of regulatory regimes, while gaining access to multiple jurisdictions. And jurisdictions retain their sovereign authority to decide what their own regulatory goals and requirements will be—choosing the regulators that satisfy them. This enables individual countries to continue to reflect their own values, culture, and risk preferences in a given domain.

We think that this reframing of harmonization efforts—away from securing global agreement on specific legal standards and towards a novel mechanism to reduce cross-jurisdictional compliance burdens—is especially important in the context of the geopolitical challenges facing AI governance. The AI race between the United States and China, with their fundamentally different approaches to industrial policy and regulation, makes the prospect for global agreement on a shared set of AI standards seem dim.[205] But a regulatory markets approach does not depend on the United States and China agreeing on specific technical standards. Each is free to license only those private regulators that achieve the outcomes they have set for their own citizens.

To be sure, countries will not make their choices about which regulators to license in an unconstrained way. Global standards could emerge as a condition of access to global markets, much as global requirements can be imposed as conditions for countries to become members of the World Trade Organization (WTO). When China joined the WTO in 2001, it was required to make changes in how state-owned enterprises were managed, increase intellectual property protections, and improve transparency and rule-of-law procedures.[206] Outside of formal global processes, continuing the abstract example from above, Country B could face pressures to give up some sovereignty *de facto* under lobbying from Provider Z, who wishes to gain access without taking on a second regula-tor. But B's decision to align with other countries would come without conceding sovereignty *de jure*—as formal harmonization efforts require.[207] Indeed, if Country B is a less wealthy country, it could profitably choose to free ride on the oversight efforts of Country A and Country B—again, without giving up sovereign authority to change its regulatory stance under political pressure in the future. And it can benefit from Provider Z's ability to generate data from its regulation by Provider 7 to allow Country B to determine that its demographic goals are, in fact, met by 7's qualitative methods.

---

205. Efforts to achieve global consensus on AI standards are, however, underway, with the global AI Safety summits, such as the ones held in the United Kingdom in November 2023 and in South Korea in May 2024. *AI Safety Summits*, FUTURE LIFE, https://futureoflife.org/project/ai-safety-summits/ [perma.cc/JYK7-WDHT]. Additionally, international dialogues on AI safety have produced agreement between Chinese and Western academics on proposed red lines for AI development in March 2024. INT'L DIALOGUES ON AI SAFETY, https://idais.ai/ [https://perma.cc/T79R-68D9].

206. Whether these changes have been durable is subject to ongoing debate. *See* Yeling Tan, *How the WTO Changed China: The Mixed Legacy of Economic Engagement*, FOREIGN AFFS., Mar.–Apr. 2021, at 90, 91–92.

A global regulatory market provides the added benefit of scale, allowing more regulators to operate and compete with different regulatory technologies and approaches. This generates competitive benefits as more regulators have greater incentives to invest in regulatory innovation to secure market share, with reduced likelihood of monopolization. A global market also generates spillover benefits, as smaller or less wealthy countries can benefit from the investments made in regulatory innovation by regulators seeking to secure licensing approval in larger or wealthier jurisdictions.

Finally, a core benefit of regulatory markets is that they promise to recruit ground level "intelligence" to regulatory innovation, and this can help us avoid the trap of focusing only on a handful of politically salient risks (like algorithmic discrimination) or being blinded by existing harms-based regulatory frameworks such as product safety regulation. As emphasized by Hadfield,[208] market mechanisms enable more information embedded in actual practices and experiences to make their way into creative regulatory problem-solving. By creating opportunities for entrepreneurs to build and profit from novel regulatory technologies and approaches, we incentivize those working most closely with AI technologies to transfer what they have learned about how to build these technologies from providers to regulators. Engineers and designers who have worked most closely with building autonomous vehicles, for example, may be best placed to develop the technologies that ensure these vehicles meet democratically established safety and traffic-management goals. Regulatory markets open up another pathway for knowledge to make its way from target to regulator.

This is a vital pathway as AI technologies continue to develop because it will identify the ways in which regulatory goals *across the board* are disrupted by these technologies. The heavy emphasis on the new harms of AI as the basic framework for designing AI governance has obscured the ways in which AI is likely to disrupt many if not most of our *existing* regulatory goals such as stable financial systems, competitive consumer goods markets, and safe and effective health care.[209] The scientists, lawyers, and compliance personnel who work in medical device companies, for example, along with the regulators who currently evaluate medical devices using conventional methods, have a front-row seat to how AI disrupts regulation in specific contexts and are attuned to specific goals. They are perhaps the actors most likely to come up with novel ideas for how to translate regulatory goals into AI-driven contexts—how to demonstrate the achievement of safety and efficacy goals, for example, using simulated data rather than human clinical trials, or how to update the testing required to accommodate the evolution of a device that continues to learn and update its software after it has been put into service. To be sure, regulated entities currently have an incentive to educate their (government) regulators about how AI disrupts the regulatory scheme, and how regulation might adapt. But existing methods of regulatory innovation—as demonstrated by the overwhelming emphasis in the AI governance debate not on regulatory disruption but rather on a list of specific AI harms—clearly are sluggish to respond. Moreover, government regulators are understandably cautious about being lobbied by their regulated entities to change regulatory approaches. A market mechanism opens the field of innovation up to a broader range of participants, beyond regulated entities themselves, and recruits the support of human and financial capital that is more diverse and capable of bearing risk.[210] Finally, a market opportunity to solve a regulatory problem creates an incentive to "see" regulatory disruption even when regulated entities or government regulators may wish to avert their eyes.

---

207. *Cf.* Ralph C. Bryant, *Brexit: Make Hard Choices but Don't Confuse Sovereignty with Autonomy*, BROOKINGS INST. (Dec. 21, 2018), https://www.brookings.edu/articles/brexit-make-hard-choices-but-dont-confuse-sovereignty-with-autonomy/ [perma.cc/H9K4-P2C5].

208. HADFIELD, *supra* note 9.

209. JAMIE AMARAT SANDHU, NOAM KOLT & GILLIAN K. HADFIELD, REGULATORY TRANSFORMATION IN THE AGE OF AI (2023), https://cifar.ca/wp-content/uploads/2023/11/CIFAR-Regulatory-Transformation-in-the-Age-of-AI.pdf [https://perma.cc/2W6H-5SUY].

### B. Limitations and Risks: Independence, Capture, and Lax Oversight

Regulatory markets come with risks around the challenge of ensuring that private regulators are competitive and independent of the entities they regulate.

Competition might fail because there is insufficient scale to support multiple regulators. If there are only two or three companies involved in developing a particular type of AI, it will be difficult to sustain a competitive market of regulators, each of which needs to regulate multiple entities and each of which needs to be at risk of losing market share to prompt continual investment in better regulatory technology. Even where there is sufficient scale, competition might not emerge if a single regulator gains too much market share or if the costs of switching regulators is too high. Competition might also fail if regulators collude. Some of these concerns can be addressed through the design of the regulatory environment imposed by governments: Antitrust and competition law could protect against the monopolization of the regulatory market, but robust competition might require additional protections such as limitations on market share or rules to reduce switching costs.

The independence of regulators will require close attention. Regulatory capture is a known risk in existing government-led regulation—both through explicit corruption and through more subtle mechanisms: campaign finance and

210. *See* HADFIELD, *supra* note 9.

lobbying, overlapping industry and regulator professional networks, the dependence of regulators on information supplied by industry, and so forth.[211] Regulatory markets put an additional layer between governments and industry. This creates a risk that private regulators, which are trying to sell their services to AI companies, will collaborate with those companies to cheat on government goals. Protecting the integrity of regulation will require governments to monitor the results achieved by private regulators and to have effective threats to condition, suspend, or revoke the licenses of regulators that skimp on performance to win the business of targets. This is a transformation of the existing problem of regulation: Regulation is only as good as the capacity and willingness of governments to regulate. With good design it is possible that regulatory markets make it easier for governments to regulate: Multiple regulators means multiple sources of data and industry expertise. Participants in the regulatory market will also have an incentive to monitor the performance of their competitors, perhaps exposing cases in which competitors are "cheating" on regulatory outcomes to achieve markets share. Further, instead of regulating, for example, 1,000 companies engaged in the production of AI systems in a given domain, government will be focused on regulating perhaps five or ten regulators. But regulatory markets only work if governments are willing to regulate private regulators. The model cannot fix a lack of political will to regulate the technology industry or concerns that countries like the United States, with a decisive lead in AI development globally, will be unwilling or unable to curtail innovation.

This concern about a lack of political will to regulate speaks to a core risk: The risk that governments will fail to invest in oversight. Two recent failures of regulatory intermediaries are vivid cautionary tales that demonstrate this. The first is the failure of credit rating agencies in the 2008 financial crisis.[212] The second is the insufficiency of FAA oversight practices in the context of the Boeing 737-MAX tragedies.[213] In the case of the credit rating agencies, these regulatory intermediaries were deliberately shielded from liability for errors in their ratings and there was no formal oversight from government.[214] In the case of Boeing, FAA oversight was grossly underfunded and inadequate, as repeated reports from government inspectors before the crashes made clear.[215] This emphasizes the need for a sustainable funding model that accords with the true cost of regulation. Under our proposal, at least some of this cost would be priced in the market as the cost of regulatory services, rather than being entirely dependent on taxation and government budgets. But it will still be the case that government oversight of private regulators will require robust funding. The regulatory markets model might reduce the amount of funding, relative to what is needed for robust regulation implemented exclusively by government actors. But it cannot eliminate the political budgeting challenge.

---

211. *See* Ernesto Dal Bó, *Regulatory Capture: A Review*, 22 OXFORD REV. ECON. POL'Y 203 (2006).

212. Frank Partnoy, *What's (Still) Wrong with Credit Ratings*, 92 WASH. L. REV. 1407, 1412–13 (2017). For fuller discussion, see Clark & Hadfield, *supra* note 7 (an earlier version of this Article).

213. Joseph Herkert et al., *The Boeing 737 MAX: Lessons for Engineering Ethics*, 26 SCI. ENG'G ETHICS. 2957, 2962–64 (2020). For fuller discussion, see Clark & Hadfield, *supra* note 7 (an earlier version of this Article).

214. Partnoy et al., *supra* note 212.

215. Herkert et al., *supra* note 213.

Relatedly, although a key motivation of the model is to address the technical deficit facing governments as they attempt to regulate AI, government oversight of private regulators will still require governments to increase their own access to in-house technical expertise. Thus, the challenge of recruiting technical experts into government, while it might be mitigated, will not be eliminated.

The converse risk to inadequate government oversight is the risk that governments will come under political pressure to displace private regulators in response to high-profile accidents or crises. Legislators might then encroach on the domain of the private regulator—moving away from specifying and evaluating outcome metrics and principles to dictating more of the details of regulation. If this is anticipated, it could undermine confidence in the power of the private regulator and diminish the willingness of targets to cooperate with them.

There is also a risk that the introduction of new actors and novel processes into the regulatory landscape could increase the complexity and opacity of regulation—with the resulting risks that bad actors could exploit or 'hack' the system more easily. Concerns about the loss of transparency with the development of regulatory technologies in the financial sector have already been voiced.[216] The model we are proposing is a fundamental transformation in the accountability structure and mechanisms for regulation—and we should expect that it will take time and thoughtful design to both identify and mitigate the new risks it creates. This is also a lesson from the Boeing disaster: Extensive reliance on companies certifying their own complex technical systems eventually revealed substantial risks of criminal conduct on the part of the target company, with deadly consequences.[217] Our model seeks to respond to these risks of regulatory evasion by introducing an independent sector of technologically sophisticated regulatory intermediaries—but we caution that powerful incentives to evade regulation will still pervade throughout the system.

Indeed, nothing in the regulatory markets model eliminates all the usual challenges in building and implementing effective regulation. Particularly in the context of intense geopolitical competition in AI technology, governments will still face powerful interests lobbying against regulation and appealing to the need to remain innovative and competitive. And as more powerful AI models proliferate across the internet, with both uncontrolled API access and open sourcing of model weights, governments will struggle to impose regulation on bad actors, foreign and domestic. These are challenges that require additional

216. Bamberger, *supra* note 180, at 735.

217. *See* Eileen Sullivan & Danielle Kaye, *Boeing Agrees to Plead Guilty to Felony in Deal with Justice Department*, N.Y. TIMES (July 8, 2024), https://www.nytimes.com/2024/07/08/business/boeing-justice-department-plea-deal.html [https://web.archive.org/web/20250606230620/https://www.nytimes.com/2024/07/08/business/boeing-justice-department-plea-deal.html].

tools;[218] regulatory markets are not a magic bullet for the regulatory state. The model will introduce new, ordinary, regulatory challenges such as ensuring that competition for regulatory licenses is not distorted by corruption. But we believe the opportunities presented by regulatory markets outweigh, at least in some potential applications, the risks so long as the risks are appropriately managed.

## V. THE URGENCY OF AI GOVERNANCE

We began with the speed and scale of AI development, and we will end with speed and scale as well. Regulatory science is not a fast-moving field. The "new governance" approaches we have highlighted have been under discussion for decades; so too has the recognition that regulatory intermediaries play an essential role in complex, modern economies. We offer up the proposal for regulatory markets however not merely as an addition to the literature on regulation. We feel considerable urgency for the regulatory community to recognize the impending and potentially widespread failure of our regulatory models to manage the enormous transformations that AI is producing. Regulatory theorists appreciate more than most just how subtle, complex, and essential regulation is in the modern world. They avoid the superficial contests in political discourse about regulation as a drag on innovation; they recognize the sense in which there is no such thing as an unregulated market.[219]

There is the risk that even if a regulatory markets solution would generate benefits in terms of improved AI safety and governance; however, it is simply infeasible to affect such a wholesale shift in regulatory methods, particularly in the current geopolitical climate. We have stressed the continuity in this proposal with existing trends in regulation: the increased use of private vendor-developed compliance technologies and greater reliance by government regulators on digital tools in financial regulation, the development of new regulatory models based on outcome metrics rather than prescriptive technological requirements, and the growing role for a network of third-party regulatory intermediaries. But at the same time our proposal goes beyond existing models. To implement regulatory markets would require governments to develop novel licensing systems. This could require new government agencies; it at least requires new regulatory metrics and methods to be designed and implemented, focused not on the ultimate requirements imposed on regulatory targets but on regulatory intermediaries. It requires a nascent set of for-profit and not-for-profit regulatory intermediaries to attract the interest of investors and begin building out more robust approaches to AI regulation in a variety of specific domains. It requires legislators to agree on legal frameworks to generate incentives for target companies to buy the services of private regulators. It requires new private regulators

218. *See, e.g.*, Gillian Hadfield et al., *It's Time to Create a National Registry for Large AI Models*, CARNEGIE ENDOWMENT FOR INT'L PEACE (July 12, 2023), https://carnegieendowment.org/2023/07/12/it-s-time-to-create-national-registry-for-large-ai-models-pub-90180#:~:text=Developing%20an%20AI%20registry%20will,or%20enable%20people%20to%20do [https://perma.cc/GHN8-EKTC].

219. HADFIELD, *supra* note 9, ch. 4.

and target companies to work together to implement novel approaches to regulation and legal liability. Above all, a shift to regulatory markets requires countries to bet on a new approach to regulation at a time when the stakes seem incredibly high: Will the United States, for example, risk regulatory missteps that could diminish its lead over China in AI development?

We think that all of the theoretical and practical challenges to the implementation of a regulatory markets model are real. But we also think that the AI governance challenge cannot be met without regulatory innovation. The urgency of the moment is that AI technologies have the capacity to rewrite the fundamental structure of human societies and economies—and they are doing so in ways today that are galloping past the regulatory frameworks put in place to manage the twentieth-century world. We think it is time for bold new ideas for regulation, and we think regulatory markets are one such idea. We believe they are a critical tool that governments can, and should, begin to deploy to meet the core governance challenges of AI.